\documentclass{article}

\usepackage{microtype}
\usepackage{graphicx}
\usepackage{subfigure}
\usepackage{booktabs} 

\usepackage{hyperref}

\usepackage[accepted]{icml2023}

\usepackage{amsmath}
\usepackage{amssymb}
\usepackage{mathtools}
\usepackage{amsthm}
\usepackage{multirow}
\usepackage{soul}
\usepackage{xcolor}
\definecolor{myblue}{RGB}{204,229,255}

\usepackage[capitalize,noabbrev]{cleveref}
\Crefname{equation}{Eq.}{Eqs.}

\usepackage[nice]{nicefrac}
\theoremstyle{plain}

\theoremstyle{definition}

\theoremstyle{remark}

\DeclareMathOperator*{\argmax}{arg\,max}
\usepackage[noabbrev]{cleveref}
\usepackage[textsize=tiny]{todonotes}
\newcommand{\KL}{\mathbb{D}_{\mathrm{KL}}}

\icmltitlerunning{Diverse and Faithful Knowledge-Grounded Dialogue Generation via Sequential Posterior Inference}

\begin{document}

\twocolumn[
\icmltitle{Diverse and Faithful Knowledge-Grounded Dialogue Generation via \\Sequential Posterior Inference}




\icmlsetsymbol{equal}{$\dagger$}
\icmlsetsymbol{equal2}{$\ddagger$}
\begin{icmlauthorlist}
\icmlauthor{Yan Xu}{xxx,equal}
\icmlauthor{Deqian Kong}{yyy,equal}
\icmlauthor{Dehong Xu}{yyy}
\icmlauthor{Ziwei Ji}{xxx}
\icmlauthor{Bo Pang}{zzz}
\icmlauthor{Pascale Fung}{xxx,equal2}
\icmlauthor{Ying Nian Wu}{yyy,equal2}
\end{icmlauthorlist}

\icmlaffiliation{xxx}{Center for Artificial Intelligence Research (CAiRE), The Hong Kong University of Science and Technology, Hong Kong}
\icmlaffiliation{yyy}{Department of Statistics, UCLA, CA, USA}
\icmlaffiliation{zzz}{Salesforce Research}

\icmlcorrespondingauthor{Yan Xu}{yxucb@connect.ust.hk}
\icmlcorrespondingauthor{Deqian Kong}{deqiankong@ucla.edu}

\icmlkeywords{posterior inference, knowledge-grounded dialogue}

\vskip 0.3in
]



\printAffiliationsAndNotice{\icmlEqualContribution\icmlEqualAdvising} 
\begin{abstract}
The capability to generate responses with diversity and faithfulness using factual knowledge is paramount for creating a human-like, trustworthy dialogue system. 
Common strategies either adopt a two-step paradigm, which optimizes knowledge selection and response generation separately, and may overlook the inherent correlation between these two tasks, or leverage conditional variational method to jointly optimize knowledge selection and response generation by employing an inference network.
In this paper, we present an end-to-end learning framework, termed \textit{Sequential Posterior Inference} (SPI), capable of selecting knowledge and generating dialogues by approximately sampling from the posterior distribution. Unlike other methods, SPI does not require the inference network or assume a simple geometry of the posterior distribution. This straightforward and intuitive inference procedure of SPI directly queries the response generation model, allowing for accurate knowledge selection and generation of faithful responses.
In addition to modeling contributions, our experimental results on two common dialogue datasets (Wizard of Wikipedia and Holl-E) demonstrate that SPI outperforms previous strong baselines according to both automatic and human evaluation metrics. The code and checkpoints are available at \url{https://github.com/deqiankong/SPI}.
\end{abstract}

\section{Introduction}
\label{sec:intro}

Open-domain dialogue systems aim at fulfilling human-machine conversations by producing human-like responses to utterances from humans~\cite{serban2016building}. 
The emergence of large-scale pre-trained language models (PLMs) has turbocharged the development of open-domain dialogue systems~\cite{zhang2020dialogpt,roller2021recipes}. By maximizing the token-level likelihood of gold responses given dialogue history, dialogue systems can generate fluent and natural responses. However, challenges remain to ensure that responses are diverse and informative~\cite{ghazvininejad2018knowledge}, yet remain factual and accurate~\cite{shuster2021retrieval}. Prior approaches for improving the diversity of dialogue responses focus on preventing them from being dull and repetitive~\cite{zhao2019low,xu2022retrieval}, while optimizing for diversity alone tends to encourage the dialogue system to hallucinate non-factual responses~\cite{ji2022survey}. ChatGPT~\cite{openai_2023} tries to address this issue using a reward model trained with human preference. However, it is very resource-consuming.
To address this limitation in generative dialogue systems, we need to ground system responses on external knowledge effectively. 

Knowledge-grounded dialogue (KGD) has been investigated in recent years~\cite{dinan2018wizard,li2020zero,xu2021caire,yang2022take}. The objective is to enhance dialogue response generation to facilitate engaging and in-depth conversations, while avoiding the inclusion of non-factual information. The task can be achieved following a two-step paradigm: (1) knowledge selection; (2) response generation. 
Some previous works~\cite{lianlearning,kim2019sequential,chen2020bridging} optimize these two steps individually. They first utilize variational inference~\cite{kingma13auto} for knowledge selection, where the prior distribution is conditioned on dialogue history, and the posterior distribution depends on both response and dialogue history. Then they optimize the response generation task based on the selected knowledge. 
Since knowledge selection in KGD tasks is a complex one-to-many problem, it is not trivial to generate a factual response with dialogue history and selected knowledge solely, not to mention that inaccurate knowledge may be chosen even with a complex knowledge selection module. Other works~\cite{liu2021three} bypass the knowledge selection step by providing all the knowledge candidates to the response generator, which is computationally inefficient.
Therefore, it is natural to choose a probabilistic model with two latent variables to select knowledge and generate responses so that both procedures can be optimized simultaneously. CoLV~\cite{zhan2021colv} follows this scheme and chooses to optimize these latent variables by recruiting an inference network to infer the posterior distribution.
However, such methods using variational inference trained with evidence lower bound (ELBO) may ignore the fact that knowledge selection is inherently correlated to response generation. Hence, there might be a large amortization gap between log-likelihood and ELBO~\cite{cremer2018inference}. 
An alternative to variational inference is  posterior inference, such as Markov Chain Monte Carlo (MCMC) which may be in the form of Langevin dynamics~\cite{langevin1908theory}. \cite{pang2021generative} proposes to generate text using short-run inference dynamics, such as finite step Langevin dynamics guided by the posterior distribution of the latent variable. Posterior inference has demonstrated its simplicity and superiority in image modeling, trajectory prediction, etc.~\cite{bo_latent, Pang_2021_CVPR, xie2022tale, li2022learning}. However, posterior inference-based methods are still under-explored in the scenarios of PLMs.

In this work, we propose a probabilistic model with dual latent variables, a discrete latent variable for knowledge selection, and a continuous latent variable for response generation. Instead of variational inference, we propose a new approximate sampling method, \textit{Sequential Posterior Inference} (SPI). This model can be learned by approximate maximum likelihood estimation (MLE). Compared to variational inference, SPI has the advantage of fewer model parameters since there is no need to parameterize the inference network, which eases the effort of fine-tuning in PLMs. To amplify the efficiency of SPI within PLMs, we propose to leverage the initializer or learnable prior to sample the discrete latent variable, and short-run MCMC to sample the continuous latent variable.
Empirically, we show that the model trained with SPI can generate faithful and diverse responses with external knowledge. Our model outperforms previous methods on both WoW and Holl-E benchmarks. Further human evaluation has demonstrated its superiority as well. 

Our contributions are three-fold:

\noindent(1) We propose a probabilistic dialogue system for KGD that can be learned by approximate MLE with sequential posterior inference (SPI).

\noindent(2) We propose to use an initializer and short-run MCMC to explore the discrete and continuous search spaces, which enables efficient approximate MLE learning in PLMs.

\noindent(3) Our proposed model achieves state-of-the-art (SOTA) performance on two common KGD benchmarks.

\section{Methods}
\label{sec:methods}
\begin{figure*}[!ht]
 \centering
 \includegraphics[width=\linewidth]{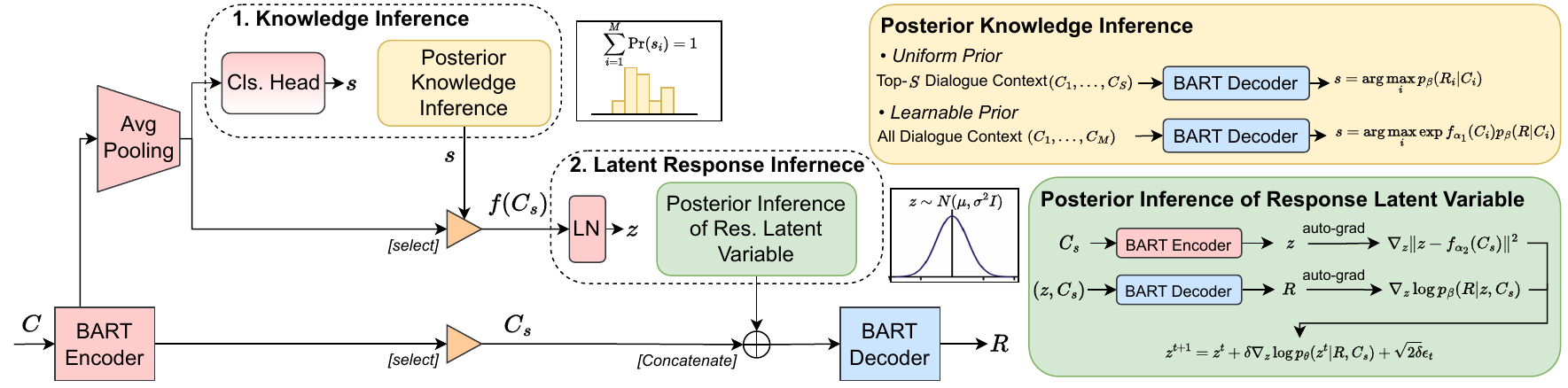}
 \vspace{-1.5em}
  \caption{The overview of the learning algorithm of \textbf{SPI} (left), where modules in pink denote the context-conditioned prior model and modules in blue denote the response generation model. The prior model is mainly instantiated with the BART encoder, while the generator is implemented with the BART decoder. We also demonstrate details of posterior knowledge selection and posterior inference of the response latent variable on the right.}
  \label{fig:system}
  \vspace{-1em}
\end{figure*}

\subsection{Model}

Suppose we have $N$ observed examples $\{\mathcal{D}^n\}_{n=1}^N$ in dialogue dataset. For each example, $\mathcal{D}^n=(C^n, R^n)$, where $C^n$ is the dialogue context, and $R^n$ is the response based on the dialogue history and selected knowledge. In KGD tasks, each dialogue context consists of dialogue history $H^n$, and a set of $M$ knowledge candidates $\mathbf{K}^n = \{K^n_i\}_{i=1}^{M}$, denoted as $C^n=(H^n, \mathbf{K}^n)$.


We consider the KGD task as a conditional generation process given dialogue history. Let $s\in\{1,\dots, M\}$ be a discrete variable indicating the choice of the knowledge candidate. Let $z\in\mathbb{R}^d$ be a $d$-dimensional continuous variable as a summary or abstraction of the future response to account for the sentence-level semantics. 
Consider the following generative model for $R$,
\begin{align}
    (s,z)\sim p_\alpha(s,z|C),\quad R\sim p_\beta(R|s,z,C),
\end{align}
where $p_\alpha(s,z|C)$ is the context-conditioned prior model parameterized by $\alpha$ and $p_\beta(R|s,z,C)$ is the response generation model parameterized by $\beta$. 

To be specific, we may factorize the context-conditioned prior model as
\begin{align}
    s\sim p_{\alpha_1}(s|C),\quad z\sim p_{\alpha_2}(z|s, C),
\end{align}
where $p_{\alpha_1}(s|C)$ can be defined as a simple uniform distribution  $\mathbb{P}(s=i)=\frac1M$, $i\in\{1,\dots,M\}$, or a learnable distribution  $\mathbb{P}(s=i)=\frac{\exp{(f_{\alpha_1}(s=i,C))}}{\sum_{s=1}^M \exp{(f_{\alpha_1}(s,C)})}$, $i \in\{1,\dots,M\}$, and $p_{\alpha_2}(z|s, C)=\mathcal{N}(f_{\alpha_2}(s,C), \mathbf{I})$ is  isotropic Gaussian. In our implementation, $f_\alpha(\cdot)$ is parameterized based on a pre-trained BART~\cite{lewis2020bart} encoder and $\alpha=(\alpha_1, \alpha_2)$ consists of the parameters of two priors.

For the response generation model, $p_\beta(R|s, z, C)$ is defined in a conditional auto-regressive manner, 
\begin{align}
    p_\beta(R|s, z, C) = \prod_{l=1}^{L} p_\beta(r_{l} |s, z, r_{<l}, C), \label{eq: generator}
\end{align}
where $L$ refers to the sentence length of the response $R$, $r_l$ is the $l$-th token of the response, and $p_\beta$ is parameterized based on a pre-trained BART decoder. Note that $s$ and $z$ control every step of the auto-regressive generation. 

The context-conditioned distribution of response $R$ is $p_\theta(R|C)=\sum_s\int p_\theta(s,z,R|C)dz$, where $\theta=\{\alpha,\beta\}$. Given $C$ and $R$, the inference of $(s,z)$ can be approximately achieved using SPI (see Section~\ref{posterior sampling}),
\begin{align}
    p_\theta(s,z|R, C) = \frac{p_\theta(s,z,R|C)}{p_\theta(R| C)}. 
\end{align}

\subsection{Learning}
Given training examples, $\{\mathcal{D}^n=(C^n, R^n)\}_{n=1}^N$, the model can be learned using maximum likelihood estimation (MLE) where the log-likelihood is 
\begin{align}
    L(\theta) = \frac{1}{N}\sum_{n=1}^{N}\log p_{\theta}(R^n|C^n), \label{eq:mle-lkhd}
\end{align}
where $\theta=\{\alpha, \beta\}$ are the learnable parameters of the model. 

Then the gradient of the log-likelihood function can be calculated by
\begin{align}
    &\quad \nabla_\theta \log p_\theta(R|C) \nonumber\\
    &= \frac{1}{p_\theta(R|C)} \nabla_\theta p_\theta(R|C)\nonumber\\
   &= \frac{1}{p_\theta(R|C)} \sum_s\int\nabla_\theta p_\theta(s,z,R|C) dz\nonumber\\
   &= \sum_s\int\frac{p_\theta(s,z,R|C)}{p_\theta(R|C)} \nabla_\theta \log p_\theta(s,z,R|C) dz\nonumber\\
   &= \mathbb{E}_{p_\theta(s,z|R, C)}[\nabla_\theta \log p_\theta(s,z,R | C)].  \label{eq:grad}
\end{align} 

Although the context-conditioned distribution $p_\theta(R|C)$ is intractable due to the latent variables being integrated out, we can approximate the above expectation using Monte Carlo samples from  the posterior $p_\theta(s, z|R, C)$ in Equation \eqref{eq:grad}, which will be further discussed in details as sequential posterior inference in \cref{posterior sampling}. 

For the gradient of the log-likelihood, we have
\begin{align}
    &\quad \nabla_\theta \log p_\theta(s,z,R | C) \nonumber\\ 
    &= \nabla_\theta \log p_\alpha(s,z|C) + \nabla_\theta \log p_\beta(R|s,z, C). \label{eq:lkhd}
\end{align}
For discrete variable $s$, if we assume a simple uniform prior distribution, $\nabla_\theta \log p_\alpha(s,z|C)=\nabla_{\theta} \log p_{\alpha_2}(z|s, C)$. Otherwise, $\nabla_\theta \log p_\alpha(s,z|C)$ becomes $\nabla_{\theta} \log p_{\alpha_1}(s|C) + \nabla_{\theta} \log p_{\alpha_2}(z|s, C)$ if we assume a learnable prior distribution.




\subsection{Sequential Posterior Inference}
\label{posterior sampling}
In Equation~\eqref{eq:grad}, the expectation can be approximated by Monte Carlo average over samples $(s, z)$ from $p_\theta(s, z|R, C)$. We define the pair between dialogue history $H$ and the $s$-th index of knowledge in $\mathbf{K}$ as $C_s=(H, K_s)$ with a slight abuse of notation.

We propose the SPI for approximate posterior inference, where we first select knowledge $p_\theta(s|R, C)$ and then infer the response latent variable~$p_\theta(z|R, C_s)$. 

\subsubsection{Posterior Knowledge Selection}
First, we shall delve into the options for posterior knowledge selection, comparing the use of a simple uniform prior with that of a learnable prior.
To infer the preferred knowledge index $s$, we shall sample from the posterior,
\begin{align}
    s\sim p_\theta(s|R, C)=\int p_\theta(s|z, R, C)p_\theta(z|R, C) dz.
    \label{eq:sample_s}
\end{align}
Since the above integration is intractable, we approximate $p_\theta(z|R, C)$ in \cref{eq:sample_s} by a point mass at the context-conditioned prior mean. Denote $\mu=f_{\alpha_2}(C_s)$. Then
\begin{align}
    s\sim p_\theta(s|z=\mu, R, C)\propto p_\theta(s, z=\mu| R, C).
    \label{eq:approx_post_s}
\end{align}
For simplicity, we still use $p_\theta(s|R, C)$ to represent the approximate posterior distribution $p_\theta(s|z=\mu, R, C)$ 
and we use $p_\beta(R|C_s)$ to represent $p_\beta(R|s, C, z=\mu)$. 

\paragraph{Uniform Prior with Top-$S$ Initializer}
For posterior inference with uniform prior, we have 
\begin{align}
    p_\theta(s|R, C) &\propto p(s|C)p_\beta(R|C_s)\nonumber\\
    &\propto p_\beta(R|C_s). 
\end{align}

In this case, the choice of the knowledge $s$ is completely dependent on the response generation model. To be concrete, for each of the knowledge candidate with its history, $C_i$, $i\in\{1,\dots,M\}$, we first concatenate it with dialogue history $H$, the posterior logits are defined by
\begin{align}
    \mathbb{P}(s=i)=\frac{p_\beta(R|C_i)}{\sum_{i=1}^M p_\beta(R|C_i)}.
\end{align}

Or we can greedily choose the one that gives the best generation performance to ease the computation,
\begin{align}
    s = \argmax_i p_\beta(R|C_i).
\end{align}

However, in the case of enormous knowledge candidates (i.e. $M$ is large), the brutal search across all $M$ candidates can be computationally inefficient. In this case, we propose to recruit an additional linear layer, $f_\gamma(C)$, (e.g., a classification head following BART encoder) as an initializer to narrow down the search space.

Based on the output logits from $f_\gamma(C)$, we can select top-$S$ knowledge candidates, where $S\ll M$. Then we can leverage the aforementioned process to select knowledge by sampling from the posterior,
\begin{align}
    \mathbb{P}(s=i)=\frac{p_\beta(R|C_i)}{\sum_{i=1}^S p_\beta(R|C_i)}.
\end{align}
Or greedily,
\begin{align}
    s = \argmax_i p_\beta(R|C_i), i \in \{1,\dots,S\}. \label{eq:infer_s}
\end{align}

This additional top-$S$ initializer can be learned using cross-entropy loss between the predicted logits and the ground-truth label. 
The selection of the ground-truth label can either be derived from gold annotations, from posterior knowledge selection, or potentially a combination of both. In our experiments, we utilize both gold annotations and selected knowledge to enhance the training of the initializer.


\paragraph{Learnable Prior}
While the fixed uniform prior is straightforward, it necessitates the use of a Top-$S$ initializer for effective operation. This leads us to contemplate if there is a way to bypass the need for an initializer, thereby enhancing the model's coherence. To this end, we employ the learnable prior, 
\begin{align}
s\sim p_{\alpha_1}(s|C) =  \frac{\exp{(f_{\alpha_1}(C_s))}}{\sum_{i=1}^M \exp{(f_{\alpha_1}(C_i)})},
\end{align}
where $f_{\alpha_1}(\cdot)$ denotes BART encoder and classification head.
For posterior distribution, we sample from,
\begin{align}
    p_\theta(s|R, C) &\propto p_{\alpha_1}(s|C)p_\beta(R|C_s),
\end{align}
or select the knowledge greedily,
\begin{align}
    s = \argmax_i \exp{(f_{\alpha_1}(C_i))} p_\beta(R|C_i), \label{eq:infer_s_learn}
\end{align}
where $i \in \{1,\dots,M\}$.

The learnable prior can be updated using either gold annotations or posterior knowledge selection. Mirroring the training of initializer, we incorporate both these elements to optimize this learnable prior.

\subsubsection{Posterior Inference of Response Latent Variable with Short-Run MCMC}
Previous work~\cite{rashkin2021increasing} defines three control codes and uses them as a prefix of the inputs to indicate how the selected knowledge is presented in the gold response. It can be considered as a high-level abstraction or summary of the future response, whereas we choose a more flexible definition of abstraction as a trainable control code or a continuous prompt that is inferred from the future response given the history and selected knowledge.

After selecting the knowledge $K_s$, we infer the continuous response latent variable $z$ by sampling from $p_\theta(z|R, C_{s})$ 
\begin{align}
z\sim p_\theta(z|R, C_s) \propto p_\theta
(z, s|R, C)
\end{align}
using Langevin dynamics
\begin{align}
    z^{t+1}=z^t + \delta \nabla_z \log p_\theta(z^t|R, C_s) + \sqrt{2\delta}\epsilon_t, \label{equ:langevin}
\end{align}
where $\epsilon_t \sim \mathcal{N}(0, \mathbf I)$, $t$ indexes the time step of the Langevin dynamics and $\delta$ is the discretization step size. The gradient term is tractable since 
\begin{align}
&\quad \nabla_z \log p_\theta(z|R, C_s) \nonumber
= \nabla_z \log p_\theta(z, R|C_{s})\nonumber\\
&= \nabla_z \log p_{\alpha_2}(z|C_{s}) + \nabla_z \log p_\beta(R|z, C_{s}), 
\end{align}
where $\log p_{\alpha_2}(z|C_{s})=\|z - f_{\alpha_2}(C_{s})\|^2/2+\mathrm{constant}$ and the second term is the response generation model. Both derivatives are tractable and can be computed by back-propagation. 

The Langevin dynamics in Equation~\eqref{equ:langevin} involves a drift term (denoted by gradient) and a diffusion term. If $z^t\sim p_\theta(z^t|R, C_{s})$,  the drift term $\nabla_z \log p_\theta(z^t|R, C_{s})$ aims to shift the distribution of $z^t$ towards basins of high log-posterior. $p_\theta(z|R, C_{s})$ can be further recovered by smoothing with the diffusion term $\sqrt{2\delta}\epsilon_t$, which induces randomness in sampling process.

However, running sufficiently long Markov chains is computationally impractical since the back-propagation through the generation model is required in each iteration according to \cref{equ:langevin}. Earlier works~\cite{pang2021generative} adopt short-run MCMC~\cite{nijkamp2019learning} in text modeling where they propose to approximately sample from the posterior distribution with a fixed small number of steps. Here we further scale up this idea in the scenario of PLMs. That said, we propose the following sampling procedure,
\begin{align}
    &z^0\sim p_{\alpha_2}(z|C_s),\nonumber\\
    &z^{t+1}=z^t + \delta \nabla_z \log p_\theta(z^t|R, C_s) + \sqrt{2\delta}\epsilon_t, \label{equ:shortrun}
\end{align}
where $t=1,\dots,T$, and the initial state for the Markov chain is sampled from the context-conditioned prior distribution. The total length of the Markov chain is rather small (e.g. $T=5$). Further theoretical underpinnings of this approximate sampling and learning method can be found in \cref{sec:theory}.

\subsection{Algorithms}
The choice of prior for the discrete variable $s$ leads to minor variations in the learning and generation algorithms.

\paragraph{Learning with Uniform Knowledge Prior}Given learning iterations $\tau=1,\dots,T_L$, the generative model with parameters $\theta=\{\alpha=\alpha_2,\beta\}$ can be updated through
\begin{align}
    &\theta_{\tau+1}=\theta_\tau + \eta_1 \Delta\theta, \nonumber \\
    &\Delta\theta = \frac{1}{N}\sum_{n=1}^N \mathbb{E}_{p_{\theta_\tau}(s^n,z^n|R^n, C^n)}[\nabla_\theta\log p_{\alpha_2}(z^n|C_s^n) \nonumber\\
    &\quad\quad + \nabla_\theta \log p_\beta(R|z^n, C_s^n)]. \label{eq:theta_update}
\end{align}

The additional top-$S$ initializer $f_\gamma(C)$ with parameters $\gamma$ can be viewed as a multi-label classifier or multiple binary classifiers and be updated by cross-entropy loss,
\begin{align}
     \mathcal{L}_\mathrm{CE}(y, C)&=-\sum_{i=1}^M  y_i \log f_\gamma(C_i)+ (1-y_i) \log (1-f_\gamma(C_i)), \nonumber \\
     \gamma_{\tau+1}&=\gamma_\tau - \eta_2 \frac{1}{N}\sum_{n=1}^N \nabla_\gamma \mathcal{L}_\mathrm{CE}(y^n, C^n), \label{eq:gamma_update}
\end{align}
where $\{y^n\}^N_{n=1}$ denotes labels which can be obtained by posterior knowledge selection and/or annotations (if we use both posterior knowledge selection and annotations, it is possible that two of $y_i$'s equal to 1). Therefore, the learned initializer can output top-$S$ candidates that likely include the gold knowledge or the one from posterior knowledge selection.

\paragraph{Learning with Learnable Knowledge Prior}The generative model with parameters $\theta=\{\alpha=(\alpha_1, \alpha_2),\beta\}$ can be updated through
\begin{align}
    &\theta_{\tau+1}=\theta_\tau + \eta_1 \Delta\theta, \nonumber \\
    &\Delta\theta = \frac{1}{N}\sum_{n=1}^N \mathbb{E}_{p_{\theta_\tau}(s^n,z^n|R^n, C^n)}[\nabla_\theta \log p_{\alpha_1}(s^n|C^n) \nonumber\\
    &\quad\quad+ \nabla_\theta \log p_{\alpha_2}(z^n|C_s^n) + \nabla_\theta \log p_\beta(R|z^n, C_s^n)]. \label{eq:theta_update_learn}
\end{align}
The learnable knowledge prior with parameter $\alpha_1$ also functions akin to a multi-label classifier. We use both gold annotations and posterior knowledge selection as ground-truth labels, mirroring the updating of the initializer.

\paragraph{Generation} We can get a response by greedy search as summarized in~\cref{alg:alg_2}. Given dialogue context $C=(H,\mathbf K)$, we can first select the knowledge candidate with the highest logit from the initializer $f_\gamma(C)$ or learnable prior $f_{\alpha_1}(C)$ based on different prior choices,
\begin{align}
    s = \argmax_i f_\gamma(C_i), \;i \in \{1,\dots,S\}.\label{eq:gen_s}
\end{align}
\begin{align}
    s = \argmax_i f_{\alpha_1}(C_i), \;i \in \{1,\dots,M\}.\label{eq:gen_s_learn}
\end{align}
Then we use sample mean of the response latent variable $p_{\alpha_2}(z|C_s) \sim \mathcal{N}(f_{\alpha_2}(C_s), \mathbf{I})$ to generate dialogue,
\begin{align}
    \hat{R}\sim p_\beta(R|z=f_{\alpha_2}(C_s), C_s).
    \label{eq:gen_z}
\end{align}

\begin{algorithm}[H]
   \caption{Learning with Sequential Posterior Inference}
   \label{alg:alg_1}
\begin{algorithmic}
   \STATE {\bfseries input:} Observed examples $\{C^n, R^n\}_{n=1}^{N_\text{train}}$, total training epochs $T_L$, learning rate $\eta$, number of candidates $S$ in knowledge selection initializer, number of Langevin steps $T$, step size $\delta$, initial weights $\theta_0, \gamma_0$. 
   \STATE {\bfseries output:} Updated weights $\theta_{T}, \gamma_{T}$.
   \FOR{$\tau=1$ {\bfseries to} $T$}
   \STATE 1. Draw observed examples $\{C^n, R^n\}$. 
   \STATE 2. \textbf{Sequential Posterior Inference}
   \STATE \quad 2.1 \textit{Posterior Knowledge Selection}
   \STATE \quad \texttt{If uniform prior}
   \STATE \quad (a) Select knowledge candidates by top-$S$ initializer. 
   \STATE \quad (b) Infer $s$ from the top-$S$ candidates using~\eqref{eq:infer_s}. 
   \STATE \quad \texttt{If learnable prior}
   \STATE \quad Infer $s$ from the learnable prior using~\eqref{eq:infer_s_learn}. 
   \STATE \quad 2.2 \textit{Posterior Inference of Response Latent Variable}
   \STATE \quad Infer $z$ by $T$-step short-run MCMC~\eqref{equ:shortrun} with step size $\delta$.
   \STATE 3. \textbf{Update Model Parameters}
   \STATE Update $\theta$ and $\gamma$ according to \eqref{eq:theta_update},\eqref{eq:gamma_update} or \eqref{eq:theta_update_learn}. 
   \ENDFOR
\end{algorithmic}
\end{algorithm}
\begin{algorithm}[H]
   \caption{Knowledge-Grounded Response Generation}
   \label{alg:alg_2}
\begin{algorithmic}
   \STATE {\bfseries input:} Observed examples $\{C^n\}_{n=1}^{N_\text{test}}$. 
   \STATE {\bfseries output:} Response $\{\hat{R}^n\}_{n=1}^{N_\text{test}}$. 
   \FOR{$n=1$ {\bfseries to} $N_\text{test}$}
   \STATE 1. Draw test example $C^n$. 
   \STATE 2. Select $s$ using the initializer $f_\gamma(C^n)$ or learnable prior $f_{\alpha_1}(C^n)$ according to \eqref{eq:gen_s} or \eqref{eq:gen_s_learn}.
   \STATE 3. Set $z$ as the sample mean of the prior $p_{\alpha_2}(z|C_s^n)$. 
   \STATE 4. Generate $\hat{R}^n$ by decoder using \eqref{eq:gen_z}. 
   \ENDFOR
\end{algorithmic}
\end{algorithm}

\section{Experiments}

\subsection{Experiment Settings}

\paragraph{Datasets}
We conduct our experiments on two KGD datasets, Wizard of Wikipedia (WoW)~\cite{dinan2019wizard} and Holl-E~\cite{moghe-etal-2018-towards}. In WoW, dialogues are directly grounded on the knowledge sentences retrieved from Wikipedia. 22.3k dialogues with 202k turns in WoW dataset are divided into training, validation, and test subsets. Both validation and test sets consist of seen and unseen sets, where the unseen set consists of the dialogues with the unseen initial topics during the training time. For a balance between task learning and generalizability, we merge two validation sets to select the best checkpoint. The Holl-E dataset contains 9k conversations with 90k utterances about movies. Each response is obtained based on unstructured knowledge such as plots, comments, and reviews about the movie. In both datasets, the gold label for knowledge selection is provided along with each dialogue turn. More details are included in \Cref{tab:data}.

\begin{table*}[t]
\centering
\resizebox{\linewidth}{!}{%
\begin{sc}
\begin{tabular}{lcccccccccccccccc}
\toprule
\multicolumn{1}{l}{\multirow{2}{*}{\textbf{Model}}} & \multicolumn{8}{c}{WoW Seen}  & \multicolumn{8}{c}{WoW Unseen}      \\ 
\cmidrule(lr){2-9} \cmidrule(lr){10-17} 
\multicolumn{1}{c}{}  & \multicolumn{1}{c}{PPL$\downarrow$} & \multicolumn{1}{c}{B3$\uparrow$} & \multicolumn{1}{c}{B4$\uparrow$} & \multicolumn{1}{c}{R1$\uparrow$} & \multicolumn{1}{c}{R2$\uparrow$} & \multicolumn{1}{c}{Dist-1$\uparrow$} & \multicolumn{1}{c}{Dist-2$\uparrow$} & \multicolumn{1}{c}{Acc$\uparrow$} & \multicolumn{1}{c}{PPL$\downarrow$} & \multicolumn{1}{c}{B3$\uparrow$}& \multicolumn{1}{c}{B4$\uparrow$} & \multicolumn{1}{c}{R1$\uparrow$} & \multicolumn{1}{c}{R2$\uparrow$} & \multicolumn{1}{c}{Dist-1$\uparrow$} & \multicolumn{1}{c}{Dist-2$\uparrow$}  &\multicolumn{1}{c}{Acc$\uparrow$}\\ \midrule
  BART$_{cat}$ & 19.7 & 6.7 &4.3 &19.3 &5.1 & 7.1 & 29.9 &$-$ &24.5 &$-$ & 4.1 & 18.9 & 4.5 &5.3 &22.2 &$-$ \\
  
  BART$_{\text{SKT}}$  & 20.3 & 7.6 &4.4   &19.4    &5.4 &6.8 &30.3 &26.8 &22.3 &$-$ & 4.6  &19  &4.7 &5.2 &24.5 &18.3 \\
  BART$_{\text{FiD}}$  & \textbf{9.5} &7.9 &5.8 &20.9 &7.8 &10.4 &39.6 &$-$ &\textbf{10.5} &8.1 & 6.1  &20.9  &7.9 &6.7 &24.2 &$-$ \\
  ZRKGC  & 40.4 & 2.8 &1.8 &18.6    &2.4 &5.4 &22.5 &$-$ &41.5 &18.6 & 1.1 &18.5  &2.4 &3.4 &15.6 &$-$\\
  DRD   & 23.0 & 7.5 &5.5  &18.0    &$-$ &$-$ &$-$ &$-$ &25.6 &16.5 & 4.3  &16.5  &$-$ &$-$ &$-$ &$-$ \\
  PIPM   & 42.7 & $-$ &3.3    &19.9   &7.3  &$-$   &26.4 &27.7 &65.7 &$-$ &2.5    &17.6    &5.4  &$-$   &17.7 &19.4\\
  CoLV &39.6 & $-$ &2.9  &20.6   &7.9  &$-$   &29.7 &30.1 &54.3 &$-$ &2.1    &19.7   &6.3  &$-$   &20.1 &18.9\\
  KAT-TSLF &14.4 & 9.1 &6.7   &21.7    &7.6 &9.5 &38.3  &$-$  &15.8 & 8.3 &6.0 &20.7    &7.2 &6.7 &\textbf{26.0}  &$-$\\
  KnowledGPT &19.2 &9.5 & 7.2 &22.0  &7.9   &8.9  &36.2  &28.0 &22.3 &8.3 & 6.0 &20.5  &6.7   &6.0  &23.8  &24.0\\
  \midrule
  \textbf{SPI}-learnable & 16.1 & \textbf{10.2} & \textbf{7.7} & \textbf{22.7} & \textbf{8.8} & 10.5 & 40.0 & \textbf{36.5} & 18.4 & \textbf{9.8} & \textbf{7.4} & 21.9 & 8.3 & 6.5 & 23.1 & \textbf{34.8} \\
  \textbf{SPI}-uniform &17.1 & \textbf{10.2} &\textbf{7.7} &\textbf{22.7}   &\textbf{8.8}  &\textbf{10.8} &\textbf{40.9}  &36.2  &19.1 &9.6 &7.3  &\textbf{22.0} &\textbf{8.5} &\textbf{6.9} &{24.3}  &34.6\\
  $\quad\nicefrac{1}{2}$ Data & 18.2 & 9.7 & 7.3 & 21.8 & 8.1 & 10.6 & 40.6 & 34.3 & 20.1 & 9.2 & 6.9 & 21.1 & 7.7 & 6.5 & 23.0 & 33.2 \\
  $\quad\nicefrac{1}{4}$ Data & 18.7 & 9.3 & 6.9 & 21.6 & 7.8 & 10.1 & 39.0 & 33.6 & 20.7 & 8.9 & 6.6 & 20.9 & 7.3 & 6.3 & 23.1 & 32.5 \\
 $\quad\nicefrac{1}{8}$ Data & 20.3 & 7.9 & 5.7 & 20.2 & 6.7 & 9.4 & 35.8 & 31.4 & 22.0 & 8.1 & 6.0 & 19.6 & 6.5 & 5.8 & 20.7 & 30.6 \\
  $\quad\nicefrac{1}{16}$ Data & 22.0 & 7.0 & 4.9 & 18.7 & 5.6 & 8.9 & 34.0 & 27.5 & 23.6 & 7.2 & 5.2 & 18.5 & 5.7 & 5.7 & 20.8 & 27.0 \\
\bottomrule
\end{tabular}
\end{sc}%
}
\caption{Automatic evaluation results on WoW test sets. PPL is short for Perplexity; B3 and B4 represent BLEU-3 and BLEU-4; R1 and R2 denote Rouge-1 and Rouge-2; Dist-1 and Dist-2 denote uni-gram and bi-gram distinct metrics. Numbers of previous models are taken from~\cite{zhao2019low,li2020zero,chen2020bridging,zhan2021colv,zhao2020knowledge,liu2021three}. SPI achieves new \text{SOTA} performance on WoW tet sets.
The performance of our proposed model under the low-resource settings is shown in the last four rows.}
\label{tab:wow}
\end{table*}

\begin{table}[t]
\centering
\resizebox{\linewidth}{!}{
\begin{sc}
\begin{tabular}{@{}lcccccccc@{}}
\toprule
\multirow{2}{*}{Model} & \multicolumn{5}{c}{Oracle Performance} & \multirow{2}{*}{FeQA} & \multicolumn{2}{c}{QuestEval} \\ \cmidrule(lr){2-6} \cmidrule(l){8-9} 
 & PPL$\downarrow$ & B3 & B4 & R1 & R2 &  & RD & RF \\ \midrule
\multicolumn{9}{l}{\textit{WoW Seen}}  \\
KnowledGPT & 9.1 & 19.2 & 15.5 & 34.5 & 17.3 & 48.1 & 42.2 & 43.5 \\
\textbf{SPI}-learnable & 8.9 & 19.3 & 15.7 & 34.6 & 17.5 & 48.3 & \textbf{45.1} & \textbf{46.6} \\
\textbf{SPI}-uniform & \textbf{8.7} & \textbf{20.0} & \textbf{16.3} & \textbf{36.1} & \textbf{18.7} & \textbf{49.2} & 44.4 & 46.0 \\ \midrule
\multicolumn{9}{l}{\textit{WoW Unseen}}  \\
KnowledGPT & 9.8 & 18.3 & 14.6 & 33.8 & 16.5 & 47.4 & 41.0 & 42.2 \\
\textbf{SPI}-learnable & 9.5 & 19.2 & 15.5 & 34.0 & 17.2 & 48.1 & \textbf{44.2} & \textbf{45.7} \\
\textbf{SPI}-uniform & \textbf{9.2} & \textbf{20.1} & \textbf{16.3} & \textbf{36.0} & \textbf{18.7} & \textbf{49.6} & 44.0 & \textbf{45.7} \\ \bottomrule
\end{tabular}
\end{sc}
}
\caption{The results on automatic faithfulness metrics on WoW test sets. The proposed model, SPI, consistently outperforms KnowledGPT on all the metrics, showing its superior faithfulness.}
\label{tab:wow_faith}
\vspace{-1.5em}
\end{table}

\paragraph{Implementation Details}

An overview of the model structure of \text{SPI} is illustrated in \Cref{fig:system}. 
We implement SPI with both uniform and learnable knowledge prior distributions, denoted as SPI-uniform and SPI-learnable, respectively. Our training implementation is based on pre-trained BART-base~\cite{lewis2020bart}. In the case of SPI-uniform, the response latent prior model with parameter $\alpha_2$ is instantiated with BART encoder followed by a linear layer; the Top-$S$ initializer is instantiated with a classification head, and the response generation model with parameter $\beta$ is instantiated with BART decoder. For SPI-learnable, the learnable knowledge prior model with parameter $\alpha_1$ and the response latent prior model with parameter $\alpha_2$ share the BART encoder. However, they differ in that the learnable prior model requires an additional classification head, and the response latent prior model requires an additional linear layer. The response generation model with parameter $\beta$ is instantiated with the BART decoder.
The inferred response latent variable $z$ is concatenated with the representation of dialogue context $C_s$ from the BART encoder on the dimension of sequence length, acting as a special token or trainable control code. BART decoder generates responses conditioned on $z$ and $C_s$ through the cross-attention mechanism in each Transformer layer. 

\paragraph{Training Details}
\label{sec:training details}
We train our model with Adam optimizer with a learning rate of 1e-7 and a weight decay of 0.005. A linear scheduler is utilized to adjust the learning rate for each step. The batch size is set as 32. We train our model on NVIDIA Geforce A6000 GPU with 15 epochs and select the best checkpoint with the lowest loss on the validation set as our final model. The responses are generated using greedy search. We set $S$ as 5 for knowledge selection initialization when using uniform prior. For Langevin dynamics, the number of Langevin steps and step size are 5 and 0.1, respectively. We discuss the training time cost with Langevin dynamics in \Cref{sec:ablation:langevin}.

\subsection{Evaluation}
\paragraph{Automatic Evaluation}
To evaluate the knowledge selection performance on both datasets, we use the accuracy (Acc) score, the ratio of the test samples where selected knowledge candidates are the same as the gold annotations.
As for estimating the quality of generated responses from different models, we utilize the classical overlap-based metrics: BLEU-3 (B3), BLEU-4 (B4)~\cite{papineni2002bleu}, Rouge-1 (R1) and Rouge-2 (R2)~\cite{lin2004rouge} to measure the distance from the golden answers. Perplexity (PPL) is the exponential negative log-likelihood of the model generating gold responses. 
We use distinct scores (Dist-1 and Dist-2)~\cite{li2016diversity} to calculate the ratio of distinct uni-gram and bi-grams at the corpus level, which reflect the diversity of generated responses.

Moreover, we adopt automatic metrics, especially for evaluating the faithfulness of the generated responses, including FeQA~\cite{durmus2020feqa}, QuestEval~\cite{scialom2021questeval}, and the overlap-based performance given the oracle knowledge. FeQA and QuestEval are both question-answering-based frameworks, relying on iterations of question generation based on the generated text and question answering (QA) given the context. The QA performance's accuracy is considered equivalent to the degree of faithfulness. QuestEval has two modes: (1) reference-dependent (RD) mode assesses the generated text with ground-truth references, and (2) reference-free (RF) mode conducts the assessment when no gold reference is available.

\paragraph{Human Evaluation}
For a comprehensive evaluation, we use human evaluation to compare the generated responses from our model with those from one of the previous SOTA models, KnowledGPT~\cite{zhao2020knowledge}.\footnote{Human evaluation is conducted on Amazon Mechanical Turk (AMT) (\url{https://www.mturk.com/}).} We assess the responses quality from three aspects: \textit{Fluency}, \textit{Relevance}, and \textit{Faithfulness}.
\textit{Fluency} assesses whether the response is complete, grammatically correct, and self-consistent without repetition, while \textit{Relevance} evaluates whether the selected knowledge and the corresponding response are relevant to the dialogue history. Both fluency and relevance are assessed using A/B testing. 
We evaluate \textit{Faithfulness} using a 4-point Likert scale. 
A faithful response should be fully supported by the dialogue context of external knowledge and history and correctly convey the information in external knowledge.
50 data samples are randomly selected from each test set, and we ensure that three annotators evaluate each sample. Further details and annotator instructions are included in \Cref{appendix sec: human eval}.

\begin{table}[t]
\centering
\resizebox{1\linewidth}{!}{
\begin{sc}
\begin{tabular}{lcccccc}
\toprule
Model      & PPL$\downarrow$  & B4 & R1  & R2 & Dist-2 & Acc \\ \midrule
SKT        & 48.9  & -    & 29.8  & 23.1  & -   & 29.2     \\
DukeNet    & 42.7 & 19.2  & 32.6 & 19.6 & 28.5  & 30.4    \\
PIPM       & 39.2 & 18.3  & 30.8 & 24.0 & 27.2  & 30.7    \\
CoLV       & 34.8 & 20.3  & 32.0 & 25.8 & 29.9  & 32.7    \\
\textbf{SPI}-uniform & \textbf{12.6} & \textbf{30.7}  & \textbf{38.3} & \textbf{31.7} & \textbf{30.6}  & \textbf{38.3}   \\ \bottomrule
\end{tabular}
\end{sc}
}
\caption{Automatic evaluation results on Holl-E test set. Numbers of previous models are taken from~\cite{kim2019sequential,meng2020dukenet,chen2020bridging,zhan2021colv}. Our model outperforms all the strong baselines and achieves new SOTA performance.}
\vspace{-1em}
\label{tab:holle}
\end{table}

\subsection{Results}
\Cref{tab:wow} and \Cref{tab:holle} report automatic evaluation results of our proposed model on WoW and Holl-E test sets. We compare our model with a number of previous strong models on both datasets and highlight the best performance of each metric in bold. The baseline models are introduced in \Cref{sec:baseline}. Comparing SPI models that learn with two prior hypotheses, SPI with uniform knowledge prior shows comparable performance on the overall response generation performance and knowledge selection accuracy, but it ensures better diversity of the generated response. Our proposed method achieves new \text{SOTA} performance on both datasets. It outperforms all the previous strong baseline models on knowledge selection accuracy and overlap-based metrics, indicating a higher quality of knowledge selection and response generation.
Comparing SPI with uniform knowledge prior and KnowledGPT, our model shows an 11.4\% on Rouge-2, and 29.3\% on accuracy on the WoW test seen set. Meanwhile, the improvements on the WoW test unseen set are even larger. This proves the better \textit{generalizability} of our model. The improvements on the Holl-E dataset are at least 17\% for all the metrics except Distinct-2 (2\%). 

Furthermore, SPI models consistently outperform KnowledGPT on all the automatic faithfulness metrics in \Cref{tab:wow_faith}, showing its superior faithfulness. Our advantage over other models in distinct scores (\Cref{tab:wow} and \Cref{tab:holle}) also shows that our model tends to generate more diverse responses, especially in the seen domain. In WoW unseen set, SPI underperforms KAT-TSLF~\cite{liu2021three} on the Distinct-2 metric. KAT-TSLF proposes a BART-based model pre-trained on a large dialogue corpus with pseudo-knowledge pairs and then adapted to WoW dataset through fine-tuning. Regarding its performance on other metrics, we believe pre-training is the major contributor to diversity. Our model achieves the second-best performance on Distinct-2 with no additional pre-training step or data resource. 

PPL scores of our model are less satisfying than the deterministic models, i.e., BART-based FiD~\cite{izacard2021leveraging}. However, it is necessary to emphasize that even though there is a correlation between PPL and human evaluation to some extent, it is not directly reflecting the quality of response generation when the PPL is low because of the likelihood trap confirmed in~\cite{zhang2021trading}.~\footnote{If the PPL of the model is too low, the correlation with human judgment decreases.}

\begin{table}[t]
\centering
\resizebox{1\linewidth}{!}{
\begin{sc}
\begin{tabular}{lcccccc}
\toprule
\multicolumn{1}{c}{\multirow{2}{*}{Model}} & \multicolumn{2}{c}{Fluency} & \multicolumn{2}{c}{Relevance} & \multicolumn{2}{c}{Faithfulness} \\ \cmidrule(lr){2-3} \cmidrule(lr){4-5} \cmidrule(lr){6-7} 
\multicolumn{1}{c}{} & Seen       & Un.    & Seen        & Un.     & Seen        & Un.        \\ 
\midrule 
KnowledGPT   & 62.5\%     & 60.3\%         & 70.8\%      & 62.2\%          & 3.33        & 3.42               \\
\textbf{SPI}-uniform   & \textbf{88.7\%}     & \textbf{83.3\%}        & 79.8\%      & \textbf{74.4\%}          & \textbf{3.66}       & \textbf{3.65}             \\
\bottomrule
\end{tabular}
\end{sc}
}
\caption{Human evaluation results on WoW test sets, in terms of \textit{Fluency}, \textit{Relevance}, and \textit{Faithfulness}. \textit{Un.} is short for the unseen set. A pairwise t-test is conducted to validate the significance of the improvements, and the corresponding results in bold are significantly better than those from the baseline model ($p < 0.05$).}
\label{tab: human evaluation}
\vspace{-1.5em}
\end{table}

\begin{table*}[t]
\centering
\resizebox{0.9\linewidth}{!}{
\begin{sc}
\begin{tabular}{@{}lcccccccccccc@{}}
\toprule
\multirow{2}{*}{Top-S} & \multicolumn{6}{c}{WoW Seen} & \multicolumn{6}{c}{WoW Unseen} \\ \cmidrule(l){2-7} \cmidrule(l){8-13} 
 & B-4 & R-2 & Dist-2 & FeQA & Q.E.(RD/RF) & Acc & B-4 & R-2 & Dist-2 & FeQA & Q.E.(RD/RF) & Acc \\ \midrule
1 & 7.3 & 8.4 & 36.6 & 40.4 & 41.1/43.0 & \textbf{37.0}  & 6.9 & 7.7 & 22.5 & 39.2 & 39.9/41.8 & \textbf{34.7} \\
3 & 7.4 & 8.3 & 39.4 & 40.7 & 41.4/43.2 & 34.1 & 7.0 & 7.8 & 22.5 & 40.5 & 40.5/42.2 & 32.2 \\
5 (Ours) & \textbf{7.7} & \textbf{8.8} & 40.9 & \textbf{49.2} & \textbf{44.4/46.0} & 36.2 & \textbf{7.3} & \textbf{8.5} & 24.3 & \textbf{49.6} & \textbf{44.0/45.7} & 34.6 \\
10 & 7.2 & 8.8 & \textbf{41.1} & 48.0 & 42.4/44.2 & 36.4 & 7.3 & 8.4 & \textbf{24.4} & 47.7 & 42.3/44.0 & 34.6 \\ \bottomrule
\end{tabular}
\end{sc}
}
\caption{Ablation study on the impact of the choice of top-$S$ for posterior knowledge selection initialization on WoW test sets. \textsc{Q.E.} is short for \textsc{QuestEval}. Our final model with top-5 knowledge candidates shows a balance between diversity and overlap-based accuracy on the quality of generated responses.}
\label{tab:top-s}
\end{table*}

\begin{table*}[t]
\centering
\resizebox{0.9\linewidth}{!}{
\begin{sc}
\begin{tabular}{@{}lccccccccccc@{}}
\toprule
\multirow{2}{*}{\begin{tabular}[c]{@{}l@{}}Langevin \\ Steps\end{tabular}} & \multicolumn{5}{c}{WoW Seen} & \multicolumn{5}{c}{WoW Unseen} & \multirow{2}{*}{\begin{tabular}[c]{@{}l@{}}Tr. Time \\ (/Epoch)\end{tabular}} \\ \cmidrule(l){2-6} \cmidrule(l){7-11} 
 & B4 & R2 & Dist-2 & FeQA & Q.E.(RD/RF) & B4 & R2 & Dist-2 & FeQA & Q.E.(RD/RF) &  \\ \midrule
0 & 7.4 & 8.7 & 40.3 & 47.4 & 43.8/45.6 & 6.9 & 8.2 & 23.5 & 48.0 & 42.9/44.6 & 3.50hrs \\
1 & 7.6 & 8.7 & 40.3 & 47.9 & 44.2/45.9 & \textbf{7.4} & 8.4 & 23.1 & 47.9 & 43.5/45.1 & 3.56hrs \\
5 (Ours) & \textbf{7.7} & \textbf{8.8} & \textbf{40.9} & \textbf{49.2} & \textbf{44.4/46.0} & 7.3 & \textbf{8.5} & \textbf{24.3} & \textbf{49.6} & \textbf{44.0/45.7} & 3.68hrs \\ \bottomrule
\end{tabular}
\end{sc}
}
\caption{Ablation study on the impact of the number of Langevin steps on WoW test sets. \textsc{Q.E.} is short for \textsc{QuestEval}. We also present the training time (Tr. Time) per epoch under each setting. As the number of Langevin steps increases, the performance on the test seen set consistently improves, while the training time cost also increases slightly.}
\label{tab:ablation langevin}
\end{table*}

\Cref{tab: human evaluation} lists the human evaluation results on both test sets of WoW, comparing KnowledGPT and SPI with uniform prior in terms of \textit{Fluency}, \textit{Relevance}, and \textit{Faithfulness}. The details about how scores are calculated are stated in \Cref{appendix sec: human eval}. A pairwise individual t-test validates the significance of the advantages of our model over KnowledGPT. Our model is more likely to generate fluent responses, select more relevant knowledge, and ensure coherence to the dialogue history. According to the criteria of Faithfulness evaluation, both KnowledGPT and our model generate partially faithful responses. 
Nevertheless, our model generates significantly more faithful responses, while enhancing diversity given the Distinct scores in \Cref{tab:wow}. Moreover, a case study is also included in \Cref{sec:case}.

\subsection{Ablation Study}
\paragraph{Low-resource settings}
Our model demonstrates high training efficiency under low-resource settings. We train our model using the same hyper-parameter settings as SPI-uniform with $\nicefrac{1}{2}$, $\nicefrac{1}{4}$, $\nicefrac{1}{8}$, and $\nicefrac{1}{16}$ of data samples on WoW datasets. From \Cref{tab:wow}, with the increasing number of training data samples, the performance of all the metrics improves consistently. With only $\nicefrac{1}{4}$ data samples, our model can still perform comparably or even better than that of other strong baseline models with full data resources. We compare our performance with KAT-TSLF under low-resource settings, as shown in \Cref{tab:low resource}. SPI with uniform knowledge prior appears to drop less on the performance under $\nicefrac{1}{4}$ and $\nicefrac{1}{8}$ data settings, with much less training cost than KAT-TSLF. KAT-TSLF relies on pre-training with a large dialogue corpus to prevent the model from poor performance under the low-resource setting. Because of pre-training, KAT-TSLF shows zero-shot KGD ability and gets better diversity in some low-resource settings. However, we find no difficulty in applying SPI for pre-training.

\paragraph{Impact of top-$S$ selection}
When learning with uniform knowledge prior, the choice of $S$ is an essential hyper-parameter. To study the impact of it, we conduct experiments when $S=1/3/5/10$ with all the other settings kept the same. As the results listed in \Cref{tab:top-s},  when the initializer is only optimized on gold labels for knowledge selection without posterior knowledge selection ($S=1$), the model performs the best knowledge selection accuracy. However, with more knowledge candidates produced by the initializer, the diversity of generated responses is on the rise, whereas the best overlap-based accuracy achieves with top-5 knowledge candidates. It shows that injecting posterior information into the initializer during training improves the faithfulness (FeQA and QuestEval scores) of the generated responses. This verifies our assumption about the inherent correlation between knowledge selection and response generation. It also proves that better knowledge selection helps with better results but does not guarantee better responses because the generation can still hallucinate and deviate from the knowledge source provided. The two paradigms of KGD tasks should be optimized jointly.

\paragraph{Impact of the number of Langevin steps}
\label{sec:ablation:langevin}

In \Cref{tab:ablation langevin}, we further study the impact of the number of Langevin steps on response generation. When training these models, all the experimental settings except the number of Langevin steps are kept the same as SPI with uniform knowledge prior. When no Langevin step is taken, the response latent variable $z$ degenerates to be a deterministic representation. Posterior inference of $z$ further boosts the performance of the SPI model on overlap-based accuracies, demonstrating the effectiveness of the proposed method, especially in the unseen domain. It also improves both diversity and faithfulness by providing a high-level abstraction of the further response with the response latent variable.
Posterior inference with Langevin dynamics requires the model to use MCMC, which sequentially queries the BART decoder to obtain the gradient from the generator for updating the response latent variable $z$. One possible concern is the increasing training cost when more Langevin steps are taken. We calculate the training time per epoch for models with different Langevin steps. Posterior inference of response latent variable $z$ with Langevin steps to be five only extends the training time per epoch by 5.1\%, which does not bring much burden on the training process.

\section{Related Work}
\paragraph{Knowledge-Grounded Dialogue Generation}

KGD task has been investigated for many years~\cite{dinan2019wizard,feng2021multidoc2dial}. 
Due to one-to-many problems in knowledge selection, one line of existing work adopts variational inference-based methods, which construct a latent variable for knowledge selection and optimize it with variational inference~\cite{lianlearning,kim2019sequential,li2020zero,chen2020bridging}. Further explorations extend the formulation to two collaborative latent variables to augment response generation or enhance knowledge selection. \cite{zhan2021colv} utilizes two collaborative latent variables to model the distributions of knowledge and response simultaneously, while \cite{fu2022there} introduces two latent variables to indicate the fragment of personal memory to evoke and the knowledge candidate to select, respectively. Another line of research bypasses the knowledge selection step but relies on improving knowledge usage during response generation given all the knowledge sentences~\cite{zhao2020knowledge,liu2021three}.
Since PLMs hallucination problem~\cite{ji2022survey} leads to some of the challenges in faithfulness, we note that to reduce hallucination in KGD systems, existing work focuses on guiding the model on correct knowledge usage~\cite{rashkin2021increasing,ji2022rho} or providing dialogue models with better knowledge augmentation by improving knowledge selection performance~\cite{shuster2021retrieval}. In this work, SPI jointly improves both processes and shows a significant faithfulness advantage through automatic and human evaluation.

\paragraph{Posterior Inference} \cite{han2017alternating} proposes to learn generative image models by alternating back-propagation, which first infers the latent variable by sampling from its posterior distribution and then updates the model parameters by usual back-propagation. Our SPI shares the same insight. To sample efficiently in the continuous latent space, \cite{tieleman2008training, nijkamp2019learning} propose different versions of MCMC to learn the generative models. Specifically, short-run MCMC~\cite{nijkamp2019learning} proposes finite-step inference dynamics guided by an energy-based model. 
We further scale up this idea in the scenarios of PLMs to sample from the continuous latent space.



\section{Conclusion}
In this work, we propose a probabilistic model with dual latent variables, one discrete latent variable for knowledge selection and one continuous latent variable for response generation. This model is effectively optimized by approximate MLE with the proposed posterior inference method, SPI. Our model has demonstrated its validity and superiority with both theoretical analysis and empirical studies. Further ablation studies show that SPI can search the discrete and continuous spaces efficiently by our proposed initializer and short-run MCMC in fine-tuning PLMs. We also find that faithfulness and diversity are emergent properties that can be improved while enhancing the inherent correlation with knowledge selection and response generation and providing the generator with a high-level abstraction of the future response.
Although in this paper, we mainly focus on KGD scenarios, our proposed method, SPI, has the potential to be applied to other knowledge-intensive tasks which require reasoning ability during text generation. We leave further exploration to future work.

\section*{Acknowledgements}
Y. N. Wu was partially supported by NSF DMS-2015577. P. Fung was partially supported by the HKJCCT21EG01 of the Hong Kong Jockey Club. 
We would like to thank the five  anonymous reviewers for their constructive comments.




\clearpage
\bibliography{example_paper}

\begin{thebibliography}{46}
\providecommand{\natexlab}[1]{#1}
\providecommand{\url}[1]{\texttt{#1}}
\expandafter\ifx\csname urlstyle\endcsname\relax
  \providecommand{\doi}[1]{doi: #1}\else
  \providecommand{\doi}{doi: \begingroup \urlstyle{rm}\Url}\fi

\bibitem[Chen et~al.(2020)Chen, Meng, Li, Chen, Xu, Xu, and
  Zhou]{chen2020bridging}
Chen, X., Meng, F., Li, P., Chen, F., Xu, S., Xu, B., and Zhou, J.
\newblock Bridging the gap between prior and posterior knowledge selection for
  knowledge-grounded dialogue generation.
\newblock In \emph{Proceedings of the 2020 conference on empirical methods in
  natural language processing (EMNLP)}, pp.\  3426--3437, 2020.

\bibitem[Cremer et~al.(2018)Cremer, Li, and Duvenaud]{cremer2018inference}
Cremer, C., Li, X., and Duvenaud, D.
\newblock Inference suboptimality in variational autoencoders.
\newblock In \emph{International Conference on Machine Learning}, pp.\
  1078--1086. PMLR, 2018.

\bibitem[Dinan et~al.(2018)Dinan, Roller, Shuster, Fan, Auli, and
  Weston]{dinan2018wizard}
Dinan, E., Roller, S., Shuster, K., Fan, A., Auli, M., and Weston, J.
\newblock Wizard of wikipedia: Knowledge-powered conversational agents.
\newblock In \emph{International Conference on Learning Representations}, 2018.

\bibitem[Dinan et~al.(2019)Dinan, Roller, Shuster, Fan, Auli, and
  Weston]{dinan2019wizard}
Dinan, E., Roller, S., Shuster, K., Fan, A., Auli, M., and Weston, J.
\newblock {W}izard of {W}ikipedia: Knowledge-powered conversational agents.
\newblock In \emph{Proceedings of the International Conference on Learning
  Representations (ICLR)}, 2019.

\bibitem[Durmus et~al.(2020)Durmus, He, and Diab]{durmus2020feqa}
Durmus, E., He, H., and Diab, M.
\newblock Feqa: A question answering evaluation framework for faithfulness
  assessment in abstractive summarization.
\newblock In \emph{Proceedings of the 58th Annual Meeting of the Association
  for Computational Linguistics}, pp.\  5055--5070, 2020.

\bibitem[Feng et~al.(2021)Feng, Patel, Wan, and Joshi]{feng2021multidoc2dial}
Feng, S., Patel, S.~S., Wan, H., and Joshi, S.
\newblock Multidoc2dial: Modeling dialogues grounded in multiple documents.
\newblock In \emph{Proceedings of the 2021 Conference on Empirical Methods in
  Natural Language Processing}, pp.\  6162--6176, 2021.

\bibitem[Fu et~al.(2022)Fu, Zhao, Tao, Wen, and Yan]{fu2022there}
Fu, T., Zhao, X., Tao, C., Wen, J.-R., and Yan, R.
\newblock There are a thousand hamlets in a thousand people’s eyes: Enhancing
  knowledge-grounded dialogue with personal memory.
\newblock In \emph{Proceedings of the 60th Annual Meeting of the Association
  for Computational Linguistics (Volume 1: Long Papers)}, pp.\  3901--3913,
  2022.

\bibitem[Ghazvininejad et~al.(2018)Ghazvininejad, Brockett, Chang, Dolan, Gao,
  Yih, and Galley]{ghazvininejad2018knowledge}
Ghazvininejad, M., Brockett, C., Chang, M.-W., Dolan, B., Gao, J., Yih, W.-t.,
  and Galley, M.
\newblock A knowledge-grounded neural conversation model.
\newblock In \emph{Proceedings of the AAAI Conference on Artificial
  Intelligence}, volume~32, 2018.

\bibitem[Han et~al.(2017)Han, Lu, Zhu, and Wu]{han2017alternating}
Han, T., Lu, Y., Zhu, S.-C., and Wu, Y.~N.
\newblock Alternating back-propagation for generator network.
\newblock In \emph{Proceedings of the AAAI Conference on Artificial
  Intelligence}, volume~31, 2017.

\bibitem[Izacard \& Grave(2021)Izacard and Grave]{izacard2021leveraging}
Izacard, G. and Grave, {\'E}.
\newblock Leveraging passage retrieval with generative models for open domain
  question answering.
\newblock In \emph{Proceedings of the 16th Conference of the European Chapter
  of the Association for Computational Linguistics: Main Volume}, pp.\
  874--880, 2021.

\bibitem[Ji et~al.(2022{\natexlab{a}})Ji, Lee, Frieske, Yu, Su, Xu, Ishii,
  Bang, Madotto, and Fung]{ji2022survey}
Ji, Z., Lee, N., Frieske, R., Yu, T., Su, D., Xu, Y., Ishii, E., Bang, Y.,
  Madotto, A., and Fung, P.
\newblock Survey of hallucination in natural language generation.
\newblock \emph{ACM Computing Surveys}, 2022{\natexlab{a}}.

\bibitem[Ji et~al.(2022{\natexlab{b}})Ji, Liu, Lee, Yu, Wilie, Zeng, and
  Fung]{ji2022rho}
Ji, Z., Liu, Z., Lee, N., Yu, T., Wilie, B., Zeng, M., and Fung, P.
\newblock Rho: Reducing hallucination in open-domain dialogues with knowledge
  grounding.
\newblock \emph{arXiv preprint arXiv:2212.01588}, 2022{\natexlab{b}}.

\bibitem[Kim et~al.(2019)Kim, Ahn, and Kim]{kim2019sequential}
Kim, B., Ahn, J., and Kim, G.
\newblock Sequential latent knowledge selection for knowledge-grounded
  dialogue.
\newblock In \emph{International Conference on Learning Representations}, 2019.

\bibitem[Kingma \& Welling(2014)Kingma and Welling]{kingma13auto}
Kingma, D.~P. and Welling, M.
\newblock Auto-encoding variational bayes.
\newblock In Bengio, Y. and LeCun, Y. (eds.), \emph{2nd International
  Conference on Learning Representations, {ICLR} 2014, Banff, AB, Canada, April
  14-16, 2014, Conference Track Proceedings}, 2014.
\newblock URL \url{http://arxiv.org/abs/1312.6114}.

\bibitem[Langevin(1908)]{langevin1908theory}
Langevin, P.
\newblock \emph{On the theory of Brownian motion}.
\newblock 1908.

\bibitem[Lewis et~al.(2020)Lewis, Liu, Goyal, Ghazvininejad, Mohamed, Levy,
  Stoyanov, and Zettlemoyer]{lewis2020bart}
Lewis, M., Liu, Y., Goyal, N., Ghazvininejad, M., Mohamed, A., Levy, O.,
  Stoyanov, V., and Zettlemoyer, L.
\newblock Bart: Denoising sequence-to-sequence pre-training for natural
  language generation, translation, and comprehension.
\newblock In \emph{ACL 2020}, 2020.

\bibitem[Li \& Han(2022)Li and Han]{li2022learning}
Li, H. and Han, T.
\newblock Learning sparse latent representations for generator model.
\newblock \emph{arXiv preprint arXiv:2209.09949}, 2022.

\bibitem[Li et~al.(2016)Li, Galley, Brockett, Gao, and Dolan]{li2016diversity}
Li, J., Galley, M., Brockett, C., Gao, J., and Dolan, W.~B.
\newblock A diversity-promoting objective function for neural conversation
  models.
\newblock In \emph{Proceedings of the 2016 Conference of the North American
  Chapter of the Association for Computational Linguistics: Human Language
  Technologies}, pp.\  110--119, 2016.

\bibitem[Li et~al.(2020)Li, Xu, Wu, Zhao, Zhao, and Tao]{li2020zero}
Li, L., Xu, C., Wu, W., Zhao, Y., Zhao, X., and Tao, C.
\newblock Zero-resource knowledge-grounded dialogue generation.
\newblock \emph{Advances in Neural Information Processing Systems},
  33:\penalty0 8475--8485, 2020.

\bibitem[Lian et~al.()Lian, Xie, Wang, Peng, and Wu]{lianlearning}
Lian, R., Xie, M., Wang, F., Peng, J., and Wu, H.
\newblock Learning to select knowledge for response generation in dialog
  systems.

\bibitem[Lin(2004)]{lin2004rouge}
Lin, C.-Y.
\newblock Rouge: A package for automatic evaluation of summaries.
\newblock In \emph{Text summarization branches out}, pp.\  74--81, 2004.

\bibitem[Liu et~al.(2021)Liu, Zhao, Li, Ren, Zhang, and Yin]{liu2021three}
Liu, S., Zhao, X., Li, B., Ren, F., Zhang, L., and Yin, S.
\newblock A three-stage learning framework for low-resource knowledge-grounded
  dialogue generation.
\newblock \emph{arXiv preprint arXiv:2109.04096}, 2021.

\bibitem[Meng et~al.(2020)Meng, Ren, Chen, Sun, Ren, Tu, and
  Rijke]{meng2020dukenet}
Meng, C., Ren, P., Chen, Z., Sun, W., Ren, Z., Tu, Z., and Rijke, M.~d.
\newblock Dukenet: A dual knowledge interaction network for knowledge-grounded
  conversation.
\newblock In \emph{Proceedings of the 43rd International ACM SIGIR Conference
  on Research and Development in Information Retrieval}, pp.\  1151--1160,
  2020.

\bibitem[Moghe et~al.(2018)Moghe, Arora, Banerjee, and
  Khapra]{moghe-etal-2018-towards}
Moghe, N., Arora, S., Banerjee, S., and Khapra, M.~M.
\newblock Towards exploiting background knowledge for building conversation
  systems.
\newblock In \emph{Proceedings of the 2018 Conference on Empirical Methods in
  Natural Language Processing}, pp.\  2322--2332, Brussels, Belgium,
  October-November 2018. Association for Computational Linguistics.
\newblock \doi{10.18653/v1/D18-1255}.
\newblock URL \url{https://aclanthology.org/D18-1255}.

\bibitem[Nijkamp et~al.(2019)Nijkamp, Hill, Zhu, and Wu]{nijkamp2019learning}
Nijkamp, E., Hill, M., Zhu, S.-C., and Wu, Y.~N.
\newblock Learning non-convergent non-persistent short-run mcmc toward
  energy-based model.
\newblock \emph{Advances in Neural Information Processing Systems}, 32, 2019.

\bibitem[OpenAI(2023)]{openai_2023}
OpenAI.
\newblock Chatgpt: Optimizing language models for dialogue, Jan 2023.
\newblock URL \url{https://openai.com/blog/chatgpt/}.

\bibitem[Pang et~al.(2020)Pang, Han, Nijkamp, Zhu, and Wu]{bo_latent}
Pang, B., Han, T., Nijkamp, E., Zhu, S.-C., and Wu, Y.~N.
\newblock Learning latent space energy-based prior model.
\newblock In Larochelle, H., Ranzato, M., Hadsell, R., Balcan, M., and Lin, H.
  (eds.), \emph{Advances in Neural Information Processing Systems}, volume~33,
  pp.\  21994--22008. Curran Associates, Inc., 2020.
\newblock URL
  \url{https://proceedings.neurips.cc/paper/2020/file/fa3060edb66e6ff4507886f9912e1ab9-Paper.pdf}.

\bibitem[Pang et~al.(2021{\natexlab{a}})Pang, Nijkamp, Han, and
  Wu]{pang2021generative}
Pang, B., Nijkamp, E., Han, T., and Wu, Y.~N.
\newblock Generative text modeling through short run inference.
\newblock \emph{arXiv preprint arXiv:2106.02513}, 2021{\natexlab{a}}.

\bibitem[Pang et~al.(2021{\natexlab{b}})Pang, Zhao, Xie, and
  Wu]{Pang_2021_CVPR}
Pang, B., Zhao, T., Xie, X., and Wu, Y.~N.
\newblock Trajectory prediction with latent belief energy-based model.
\newblock In \emph{Proceedings of the IEEE/CVF Conference on Computer Vision
  and Pattern Recognition (CVPR)}, pp.\  11814--11824, June 2021{\natexlab{b}}.

\bibitem[Papineni et~al.(2002)Papineni, Roukos, Ward, and
  Zhu]{papineni2002bleu}
Papineni, K., Roukos, S., Ward, T., and Zhu, W.-J.
\newblock Bleu: a method for automatic evaluation of machine translation.
\newblock In \emph{Proceedings of the 40th annual meeting of the Association
  for Computational Linguistics}, pp.\  311--318, 2002.

\bibitem[Rashkin et~al.(2021)Rashkin, Reitter, Tomar, and
  Das]{rashkin2021increasing}
Rashkin, H., Reitter, D., Tomar, G.~S., and Das, D.
\newblock Increasing faithfulness in knowledge-grounded dialogue with
  controllable features.
\newblock In \emph{Proceedings of the 59th Annual Meeting of the Association
  for Computational Linguistics and the 11th International Joint Conference on
  Natural Language Processing (Volume 1: Long Papers)}, pp.\  704--718, 2021.

\bibitem[Robbins \& Monro(1985)Robbins and Monro]{kl}
Robbins, H. and Monro, S.
\newblock A stochastic approximation method.
\newblock In \emph{Herbert Robbins Selected Papers}, pp.\  102--109. Springer,
  1985.

\bibitem[Roller et~al.(2021)Roller, Dinan, Goyal, Ju, Williamson, Liu, Xu, Ott,
  Smith, Boureau, et~al.]{roller2021recipes}
Roller, S., Dinan, E., Goyal, N., Ju, D., Williamson, M., Liu, Y., Xu, J., Ott,
  M., Smith, E.~M., Boureau, Y.-L., et~al.
\newblock Recipes for building an open-domain chatbot.
\newblock In \emph{EACL}, 2021.

\bibitem[Scialom et~al.(2021)Scialom, Dray, Lamprier, Piwowarski, Staiano,
  Wang, and Gallinari]{scialom2021questeval}
Scialom, T., Dray, P.-A., Lamprier, S., Piwowarski, B., Staiano, J., Wang, A.,
  and Gallinari, P.
\newblock Questeval: Summarization asks for fact-based evaluation.
\newblock In \emph{Proceedings of the 2021 Conference on Empirical Methods in
  Natural Language Processing}, pp.\  6594--6604, 2021.

\bibitem[Serban et~al.(2016)Serban, Sordoni, Bengio, Courville, and
  Pineau]{serban2016building}
Serban, I., Sordoni, A., Bengio, Y., Courville, A., and Pineau, J.
\newblock Building end-to-end dialogue systems using generative hierarchical
  neural network models.
\newblock In \emph{Proceedings of the AAAI Conference on Artificial
  Intelligence}, volume~30, 2016.

\bibitem[Shuster et~al.(2021)Shuster, Poff, Chen, Kiela, and
  Weston]{shuster2021retrieval}
Shuster, K., Poff, S., Chen, M., Kiela, D., and Weston, J.
\newblock Retrieval augmentation reduces hallucination in conversation.
\newblock In \emph{Findings of the Association for Computational Linguistics:
  EMNLP 2021}, pp.\  3784--3803, 2021.

\bibitem[Tieleman(2008)]{tieleman2008training}
Tieleman, T.
\newblock Training restricted boltzmann machines using approximations to the
  likelihood gradient.
\newblock In \emph{Proceedings of the 25th international conference on Machine
  learning}, pp.\  1064--1071, 2008.

\bibitem[Xie et~al.(2022)Xie, Zhu, Li, and Li]{xie2022tale}
Xie, J., Zhu, Y., Li, J., and Li, P.
\newblock A tale of two flows: cooperative learning of langevin flow and
  normalizing flow toward energy-based model.
\newblock \emph{arXiv preprint arXiv:2205.06924}, 2022.

\bibitem[Xu et~al.(2021)Xu, Ishii, Winata, Lin, Madotto, Liu, Xu, and
  Fung]{xu2021caire}
Xu, Y., Ishii, E., Winata, G.~I., Lin, Z., Madotto, A., Liu, Z., Xu, P., and
  Fung, P.
\newblock Caire in dialdoc21: Data augmentation for information seeking
  dialogue system.
\newblock In \emph{Proceedings of the 1st Workshop on Document-grounded
  Dialogue and Conversational Question Answering (DialDoc 2021)}, pp.\  46--51,
  2021.

\bibitem[Xu et~al.(2022)Xu, Ishii, Cahyawijaya, Liu, Winata, Madotto, Su, and
  Fung]{xu2022retrieval}
Xu, Y., Ishii, E., Cahyawijaya, S., Liu, Z., Winata, G.~I., Madotto, A., Su,
  D., and Fung, P.
\newblock Retrieval-free knowledge-grounded dialogue response generation with
  adapters.
\newblock In \emph{Proceedings of the Second DialDoc Workshop on
  Document-grounded Dialogue and Conversational Question Answering}, pp.\
  93--107, 2022.

\bibitem[Yang et~al.(2022)Yang, Lin, Li, Meng, Wang, Wang, and
  Zhou]{yang2022take}
Yang, C., Lin, Z., Li, J., Meng, F., Wang, W., Wang, L., and Zhou, J.
\newblock Take: Topic-shift aware knowledge selection for dialogue generation.
\newblock In \emph{Proceedings of the 29th International Conference on
  Computational Linguistics}, pp.\  253--265, 2022.

\bibitem[Zhan et~al.(2021)Zhan, Shen, Chen, and Zhang]{zhan2021colv}
Zhan, H., Shen, L., Chen, H., and Zhang, H.
\newblock Colv: A collaborative latent variable model for knowledge-grounded
  dialogue generation.
\newblock In \emph{Proceedings of the 2021 Conference on Empirical Methods in
  Natural Language Processing}, pp.\  2250--2261, 2021.

\bibitem[Zhang et~al.(2021)Zhang, Duckworth, Ippolito, and
  Neelakantan]{zhang2021trading}
Zhang, H., Duckworth, D., Ippolito, D., and Neelakantan, A.
\newblock Trading off diversity and quality in natural language generation.
\newblock In \emph{Proceedings of the Workshop on Human Evaluation of NLP
  Systems (HumEval)}, pp.\  25--33, 2021.

\bibitem[Zhang et~al.(2020)Zhang, Sun, Galley, Chen, Brockett, Gao, Gao, Liu,
  and Dolan]{zhang2020dialogpt}
Zhang, Y., Sun, S., Galley, M., Chen, Y.-C., Brockett, C., Gao, X., Gao, J.,
  Liu, J., and Dolan, W.~B.
\newblock Dialogpt: Large-scale generative pre-training for conversational
  response generation.
\newblock In \emph{Proceedings of the 58th Annual Meeting of the Association
  for Computational Linguistics: System Demonstrations}, pp.\  270--278, 2020.

\bibitem[Zhao et~al.(2019)Zhao, Wu, Tao, Xu, Zhao, and Yan]{zhao2019low}
Zhao, X., Wu, W., Tao, C., Xu, C., Zhao, D., and Yan, R.
\newblock Low-resource knowledge-grounded dialogue generation.
\newblock In \emph{International Conference on Learning Representations}, 2019.

\bibitem[Zhao et~al.(2020)Zhao, Wu, Xu, Tao, Zhao, and Yan]{zhao2020knowledge}
Zhao, X., Wu, W., Xu, C., Tao, C., Zhao, D., and Yan, R.
\newblock Knowledge-grounded dialogue generation with pre-trained language
  models.
\newblock In \emph{Proceedings of the 2020 Conference on Empirical Methods in
  Natural Language Processing (EMNLP)}, pp.\  3377--3390, 2020.

\end{thebibliography}
\bibliographystyle{icml2023}

\newpage
\appendix
\onecolumn

\section{Theoretical Understanding}
\label{sec:theory}
In \cref{sec:methods}, we sample from $p_\theta(s, z|C, R)$ approximately. Let $q_\theta(s, z|C, R)$ be the actual distribution of the sampled $(s, z)$.
 

Given model parameters $\theta_{\tau}$ at training iteration $\tau$, the updating rule using the approximate posterior distribution of $(s,z)$ is one-step gradient ascent on the following function,
\begin{align}
    Q(\theta) &= \frac1N\sum_{n=1}^N\mathbb{E}_{q_{\theta_\tau}(s^n,z^n|R^n, C^n)}[\log p_{\theta}(s^n,z^n,R^n | C^n)]. 
\end{align}


Comparing to the log-likelihood in \cref{eq:mle-lkhd}, we have,
\begin{align}
    Q(\theta)
    &=L(\theta)+\frac1N\sum_{n=1}^N\mathbb{E}_{q_{\theta_\tau}(s^n,z^n|R^n, C^n)}[\log p_{\theta}(s^n,z^n |R^n, C^n)]\nonumber\\
    &=L(\theta)-\frac1N\sum_{n=1}^N\KL(q_{\theta_\tau}(s^n,z^n|R^n, C^n)||p_{\theta}(s^n,z^n|R^n, C^n)) \nonumber\\
    &+\frac1N\sum_{n=1}^N \mathbb{E}_{q_{\theta_\tau}(s^n,z^n|R^n, C^n)}[\log q_{\theta_\tau}(s^n,z^n|R^n, C^n)].
\end{align}

With $\theta_\tau$ fixed, the above equation becomes a function of $\theta$. Then the updating rule follows the stochastic gradient of
\begin{align}
    \Tilde{Q}(\theta)
    &=L(\theta)-\frac1N\sum_{n=1}^N\KL(q_{\theta_\tau}(s^n,z^n|R^n, C^n)||p_{\theta}(s^n,z^n|R^n, C^n)),
\end{align}
which can be viewed as a perturbation or variational lower bound of $L(\theta)$.

The fixed point of the learning algorithm that updates $\theta$ in \cref{eq:theta_update} solves the following estimating equation:
\begin{align}
    \frac{1}{N}\sum_{n=1}^N \mathbb{E}_{q_{\theta(s^n,z^n|R^n, C^n)}}\left[\nabla_\theta \log p_\theta(s^n,z^n,R^n | C^n)\right]=0. 
\end{align}

The Monte Carlo approximation of the above expectation leads to the Robbins-Monro algorithm for stochastic approximation~\cite{kl}. The convergence to the fixed point follows the regular conditions of the Robbins-Monro algorithm.

\begin{table*}[t]
\centering
\resizebox{\linewidth}{!}{%
\begin{sc}
\begin{tabular}{lcccccccccccccc}
\toprule
\multicolumn{1}{l}{\multirow{2}{*}{\textbf{Model}}} & \multicolumn{7}{c}{WoW Seen}  & \multicolumn{7}{c}{WoW Unseen}      \\ 
\cmidrule(lr){2-8} \cmidrule(lr){9-15} 
\multicolumn{1}{c}{}  & \multicolumn{1}{c}{PPL$\downarrow$} & \multicolumn{1}{c}{B3$\uparrow$} & \multicolumn{1}{c}{B4$\uparrow$} & \multicolumn{1}{c}{R1$\uparrow$} & \multicolumn{1}{c}{R2$\uparrow$} & \multicolumn{1}{c}{Dist-1$\uparrow$} & \multicolumn{1}{c}{Dist-2$\uparrow$} & \multicolumn{1}{c}{PPL$\downarrow$} & \multicolumn{1}{c}{B3$\uparrow$}& \multicolumn{1}{c}{B4$\uparrow$} & \multicolumn{1}{c}{R1$\uparrow$} & \multicolumn{1}{c}{R2$\uparrow$} & \multicolumn{1}{c}{Dist-1$\uparrow$} & \multicolumn{1}{c}{Dist-2$\uparrow$} \\ \midrule
DRD &23.0 & 7.5 &5.5 &18.0   &$-$  &$-$ &$-$   &25.6 &6.2 &4.3  &16.5 &$-$ &$-$ & $-$ \\
$\quad\nicefrac{1}{2}$ Data & 25.3 & 7.3 & 5.3 & 17.5 & $-$ & $-$ & $-$ & 27.7 & 6.4 & 4.5 & 16.7 & $-$ & $-$ & $-$ \\
$\quad\nicefrac{1}{4}$ Data & 29.2 & 6.4 & 4.4 & 16.9 & $-$ & $-$ & $-$ & 32.4 & 6.0 & 4.1 & 16.2 & $-$ & $-$ & $-$  \\
$\quad\nicefrac{1}{8}$ Data & 33.5 & 5.9 & 3.9 & 16.3 & $-$ & $-$ & $-$ & 35.8 & 5.4 & 3.5 & 16.0 & $-$ & $-$ & $-$  \\ 
$\quad\nicefrac{1}{16}$ Data & 38.6 & 5.2 & 3.3 & 15.7 & $-$ & $-$ & $-$ & 41.0 & 5.0 & 3.2 & 15.3 & $-$ & $-$ & $-$ \\ \midrule
KAT-TSLF & \textbf{14.4} & 9.1 & 6.7 & 21.7 & 7.6 & 9.5 & 38.3 & \textbf{15.8} & 8.3 & 6.0 & 20.7 & 7.2 & 6.7 & \textbf{26.0} \\
$\quad\nicefrac{1}{4}$ Data & 17.6 & 7.7 & 5.5 & 20.3 & 6.8 & 9.9 & 39.1 & 18.4 & 7.5 & 5.2 & 19.9 & 6.4 & 6.6 & 25.1 \\
$\quad\nicefrac{1}{8}$ Data & 18.8 & 7.1 & 4.9 & 19.8 & 6.3 & 9.9 & 39.5 & 20.1 & 7 & 4.8 & 19.0 & 5.9 & 6.6 & 25.3 \\
$\quad$ Zero Data & 100+ & 4.0 & 2.2 & 14.7 & 3 & 7.5 & 33.9 & 100+ & 4.7 & 2.7 & 14.9 & 3 & 5.7 & 26.4 \\ \midrule
SPI (Top-$S$, Ours) &17.1 & \textbf{10.2} &\textbf{7.7} &\textbf{22.7}   &\textbf{8.8}  &\textbf{10.8} &\textbf{40.9}   &19.1 &\textbf{9.6} &\textbf{7.3}  &\textbf{22.0} &\textbf{8.5} &\textbf{6.9} &{24.3} \\
$\quad\nicefrac{1}{2}$ Data & 18.2 & 9.7 & 7.3 & 21.8 & 8.1 & 10.6 & 40.6 & 20.1 & 9.2 & 6.9 & 21.1 & 7.7 & 6.5 & 23.0 \\
$\quad\nicefrac{1}{4}$ Data & 18.7 & 9.3 & 6.9 & 21.6 & 7.8 & 10.1 & 39.0 & 20.7 & 8.9 & 6.6 & 20.9 & 7.3 & 6.3 & 23.1  \\
$\quad\nicefrac{1}{8}$ Data & 20.3 & 7.9 & 5.7 & 20.2 & 6.7 & 9.4 & 35.8 & 22.0 & 8.1 & 6.0 & 19.6 & 6.5 & 5.8 & 20.7  \\ 
$\quad\nicefrac{1}{16}$ Data & 22.0 & 7.0 & 4.9 & 18.7 & 5.6 & 8.9 & 34.0 & 23.6 & 7.2 & 5.2 & 18.5 & 5.7 & 5.7 & 20.8 \\ \bottomrule
\end{tabular}
\end{sc}%
}
\caption{Automatic evaluation results on WoW test sets under low-resource settings, compared with DRD~\cite{zhao2019low} and KAT-TSLF~\cite{liu2021three}. PPL is short for Perplexity; B3 and B4 represent BLEU-3 and BLEU-4; R1 and R2 denote Rouge-1 and Rouge-2; Dist-1 and Dist-2 denote uni-gram and bi-gram distinct metrics.}
\label{tab:low resource}
\end{table*}

\section{Technical Challenges of Langevin Dynamics in PLMs}

The technical challenges of Langevin Dynamics in PLMs stem from the difficulties of leveraging latent variable models (LVM) in PLMs. We propose to infer the posterior distribution of discrete and continuous latent variables. 
For continuous latent variables, we mainly face two major difficulties:  (1) The choice of hyper-parameters:  step size and total number of steps. The step size determines the induced values for the drift and diffusion terms in \Cref{equ:shortrun}. The total number of steps determines the total number of times we need to back-propagate through the BART decoder (generation model) using PyTorch auto-differentiation to calculate the gradient manually. Empirically, we find that the training process is more stable when the step size is smaller than $ 0.1$ and the total number of steps is around five in our settings. (2) The initial distribution of the Markov chain is crucial. \cite{pang2021generative} uses noise-initialized Markov chains for text generation. However, we find that noise initialization does not work well in PLMs. Ideally, we should run an infinitely long chain until convergence so that the final state of the Markov Chain is independent of the initial point. In the short-run case, however, the target distribution depends on the starting point~\cite{nijkamp2019learning}. In our case, we start from the prior distribution directly. We also employ other engineering tricks, such as gradient clamping, which can be found in our released code.

\section{Baseline Models}
\label{sec:baseline}
In this work, we compare the performance of our model with nine other strong baseline models on two KGD benchmarks. We introduce each of them below:

\paragraph{SKT} SKT~\cite{kim2019sequential} is a sequential latent knowledge selection model for multi-turn KGD tasks. Both prior and posterior distributions for knowledge selection are considered sequential processes. The model can keep track of prior and posterior distributions over knowledge, where both distributions are sequentially updated considering the responses in previous turns.
We adopt the knowledge candidate selected by SKT to BART for response generation.

\paragraph{FiD} Fusion-in-Deocder (FiD)~\cite{izacard2021leveraging} is a simple yet effective model for general knowledge-intensive tasks when the context should be augmented with multiple extra documents. The model can be applied to any encoder-decoder-based PLMs. It encodes different context-document pairs in parallel and concatenates all the output hidden states together so that the decoder can attend to all these representations during generation. In this work, we use BART-base as the backbone model for a fair comparison.

\paragraph{DRD} DRD~\cite{zhao2019low} proposes a disentangled response decoder in order to isolate parameters for generating responses that depend on dialogue context, knowledge inputs, or responses themselves. When generating one token, the decoder needs to do inference with three groups of parameters respectively and decides the final output token with a decoding manager. The proposed model is also pre-trained on the same dialogue corpus as \cite{liu2021three}, thus it is also able to perform KGD generation under a low-resource setting (\Cref{tab:low resource}).

\paragraph{ZRKGC} ZRKGC~\cite{li2020zero} focuses on the situation when the real knowledge-grounded dialogue data are not available during the training process. Two latent variables that represent the knowledge for grounding and the rate of grounding are introduced to the model. The generation process is then formalized within a probabilistic framework and optimized via variational inference.

\paragraph{PIPM} PIPM is short for ``SKT+PIPM+KDBTS''~\cite{chen2020bridging}. The authors propose posterior information prediction and knowledge distillation-based training
strategy for knowledge selection. KL divergence is leveraged to bridge the gap between prior and posterior knowledge selection.

\paragraph{DukeNet} DukeNet~\cite{meng2020dukenet} explicitly models knowledge tracking and knowledge shifting and formulating their interactions as dual learning without extra external supervision.

\paragraph{CoLV} CoLV~\cite{zhan2021colv} is a Collaborative Latent Variable model. Similar to our model, it also simultaneously improves the diversity of both knowledge selection and knowledge-aware response generation. However, the model still depends on variational inference for building both latent spaces.

\paragraph{KnowledGPT} KnowledGPT~\cite{zhao2020knowledge}, as one of the previous SOTA models, equips response generation with a sequential knowledge selector and jointly optimizes both the knowledge selector and the response generator with reinforcement learning and curriculum learning. The knowledge selector first ranks all the knowledge candidates and then knowledge candidates are concatenated with dialogue history as inputs and truncated to meet the length constraint of the GPT-2.

\paragraph{KAT-TSLF} KAT-TSLF~\cite{liu2021three} is also one of the previous SOTA models. The authors propose a three-stage learning framework for low-resource knowledge-grounded dialogue tasks. First, the dialogue history encoder and knowledge encoder are pre-trained on the dialogue corpus and knowledge base respectively. Then, the method matches each dialogue turn in the dialogue corpus with a pseudo gold knowledge from the knowledge base and use the processed new corpus to pre-train the whole model. After two-stage pre-training, the model is adapted to downstream KGD benchmarks and maintains strong performance under low-resource settings. Instead of selecting the knowledge from knowledge candidates, all the provided knowledge sentences are used as inputs, and the decoder is trained to select from all the information.

\begin{table}[]
\centering
\resizebox{0.8\linewidth}{!}{%
\begin{sc}
\begin{tabular}{@{}l|cccc|ccc@{}}
\toprule
\multirow{2}{*}{}           & \multicolumn{4}{c}{Wizard of Wikipedia}  & \multicolumn{3}{|c}{Holl-E} \\ \cmidrule(l){2-5} \cmidrule(l){6-8} 
                            & Train  & Valid & Test Seen & Test Unseen & Train   & Valid   & Test   \\ \midrule
\# Conversations            & 18,430 & 1,948 & 965       & 968         & 7,228   & 930     & 913    \\
Avg. \# Knowledge Sentences & \multicolumn{4}{c}{60}                   & \multicolumn{3}{|c}{58}     \\
Avg. \# Turns               & \multicolumn{4}{c}{9}                    & \multicolumn{3}{|c}{5}      \\ \bottomrule
\end{tabular}
\end{sc}
}
\caption{Data statstics of Wizard of Wikipedia and Holl-E datasets.}
\label{tab:data}
\end{table}

\begin{table*}[t]
\centering
\begin{tabular}{@{}l|l|l@{}}
\toprule
\multicolumn{1}{l|}{} & Topic & Chevrolet Corvette \\ \midrule
\multirow{3}{*}{\begin{tabular}[c]{@{}c@{}}Dialogue \\History\end{tabular}} & User & What do you know about the Chevrolet Corvette? \\
 & System & \begin{tabular}[c]{@{}l@{}}The Chevy Corvette, or "vette" as it is known, is an iconic American sports car \\ that has been produced for half a century.\end{tabular} \\
 & User & Do you remember the prince song \colorbox{pink}{Little Red Corvette}? \\ \midrule
\multirow{2}{*}{\begin{tabular}[c]{@{}c@{}}Selected \\Knowledge\end{tabular}} & KnowledGPT & \begin{tabular}[c]{@{}l@{}}\colorbox{pink}{Chevrolet Corvette}: The first model, a convertible, was introduced at the GM \\ Motorama in 1953 as a concept show car.\end{tabular} \\
 & SPI (Ours) & \begin{tabular}[c]{@{}l@{}}Little Red Corvette: "Little Red Corvette" is a song by American musician \\ Prince.\end{tabular} \\ \midrule
\multirow{2}{*}{Response} & KnowledGPT & Yes, it was first introduced at the GM Motorama in 1953 as a concept show car. \\
 & SPI (Ours) & I do. It was a song by American musician Prince. \\ \bottomrule
\end{tabular}
\caption{One case from test seen set of WoW, comparing the generated response from SPI with that from KnowledGPT.}
\label{tab:case1}
\end{table*}

\section{Case Study}
\label{sec:case}

\Cref{tab:case1} demonstrates two typical cases from WoW test sets, comparing SPI with KnowledGPT. In the presented case, the dialogue history shows a topic shift from ``Chevrolet Corvette'' to the ``Little Red Corvette'' song. However, KnowledGPT fails to capture this shift, whereas our model selects the most relevant knowledge. 

\section{Human Evaluation}
\label{appendix sec: human eval}

\begin{figure*}[!ht]
 \centering
 \includegraphics[width=1\linewidth]{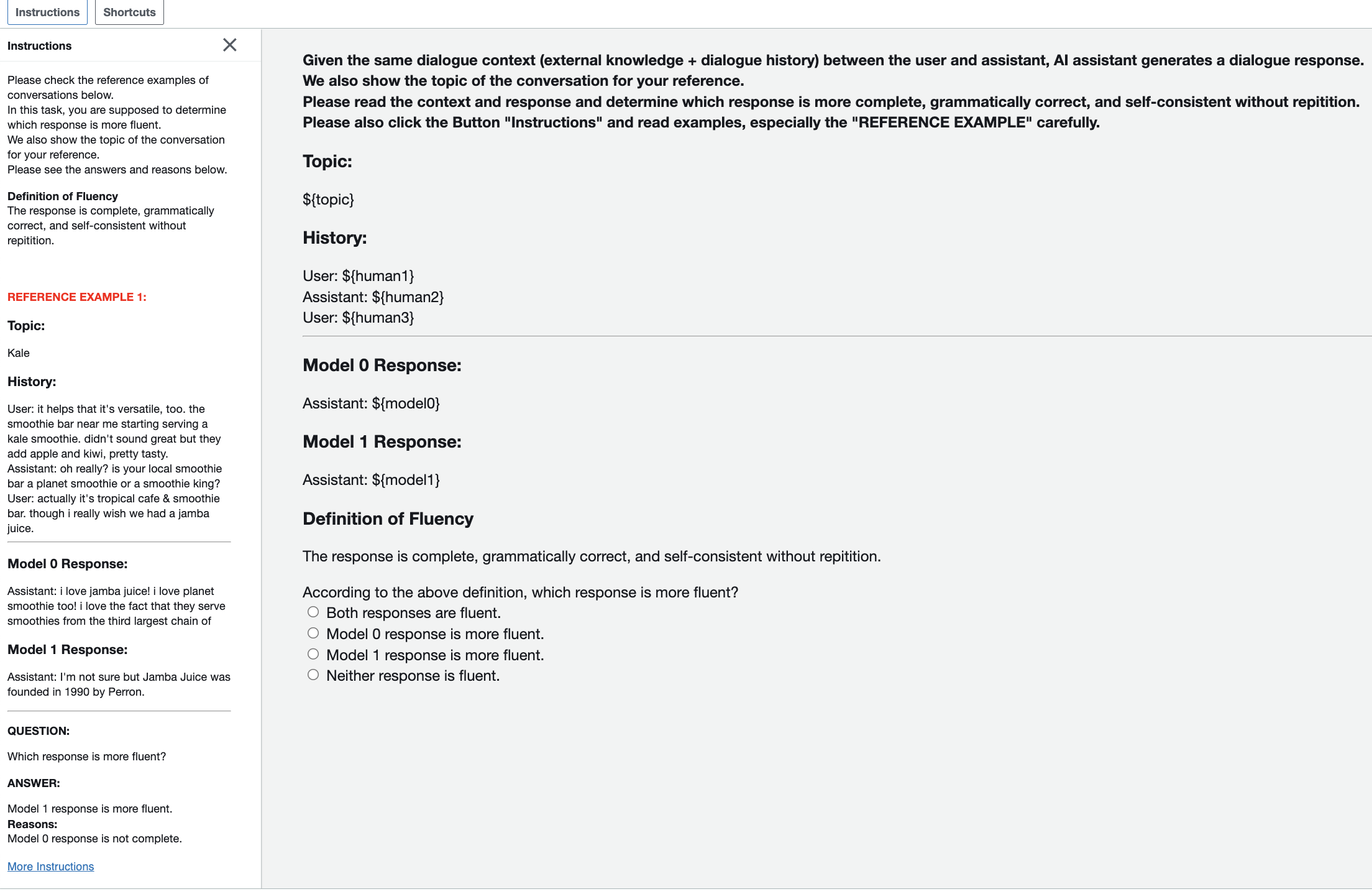}

  \caption{The annotator instruction for human evaluation on fluency via A/B testing.}
  \label{fig:human eval fluency}
\end{figure*}

\begin{figure*}[!ht]
 \centering
 \includegraphics[width=1\linewidth]{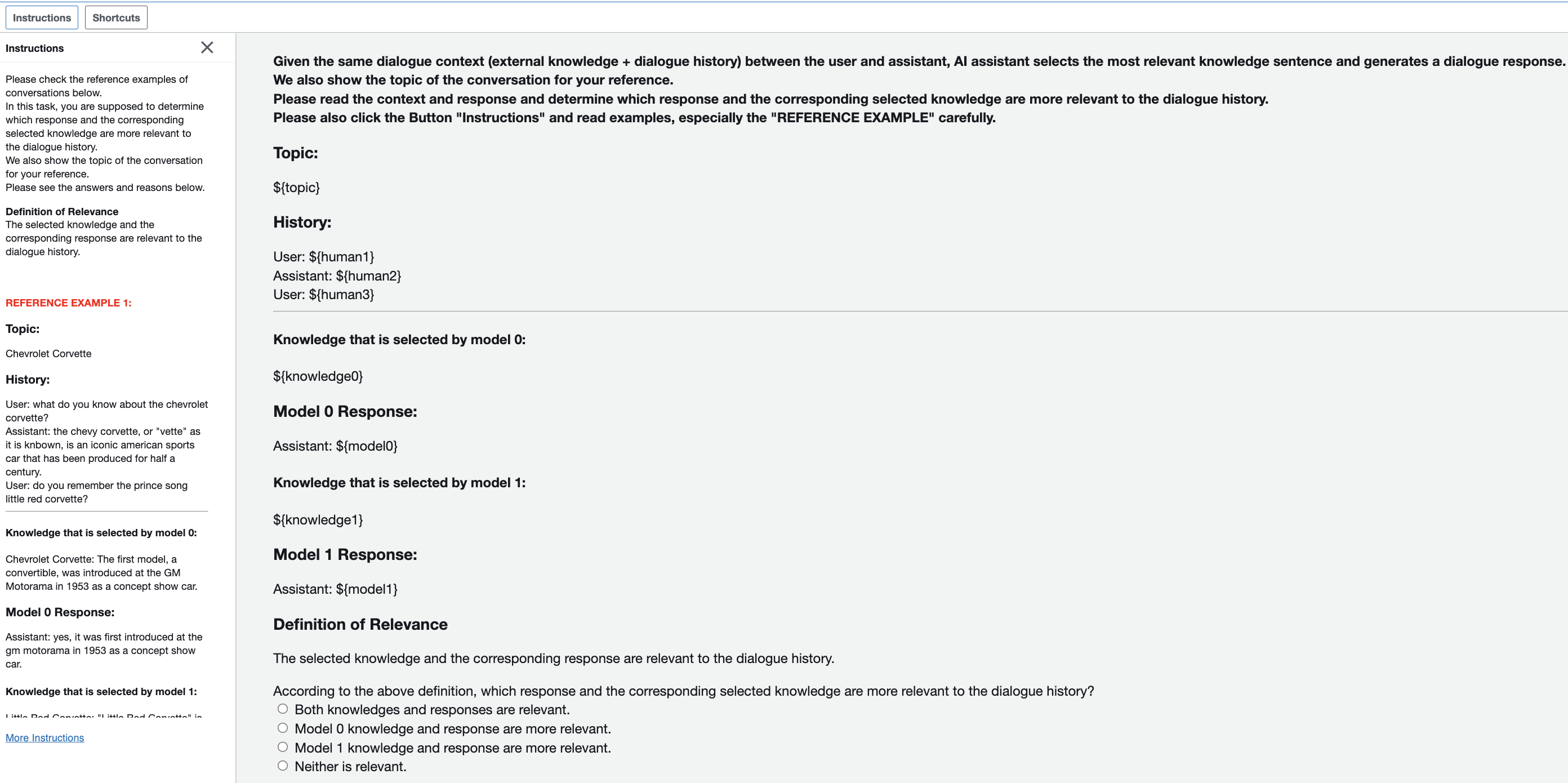}

  \caption{The annotator instructions for human evaluation on knowledge relevance via A/B testing.}
  \label{fig:human eval relevance}
\end{figure*}

\begin{figure*}[!ht]
 \centering
 \includegraphics[width=\linewidth]{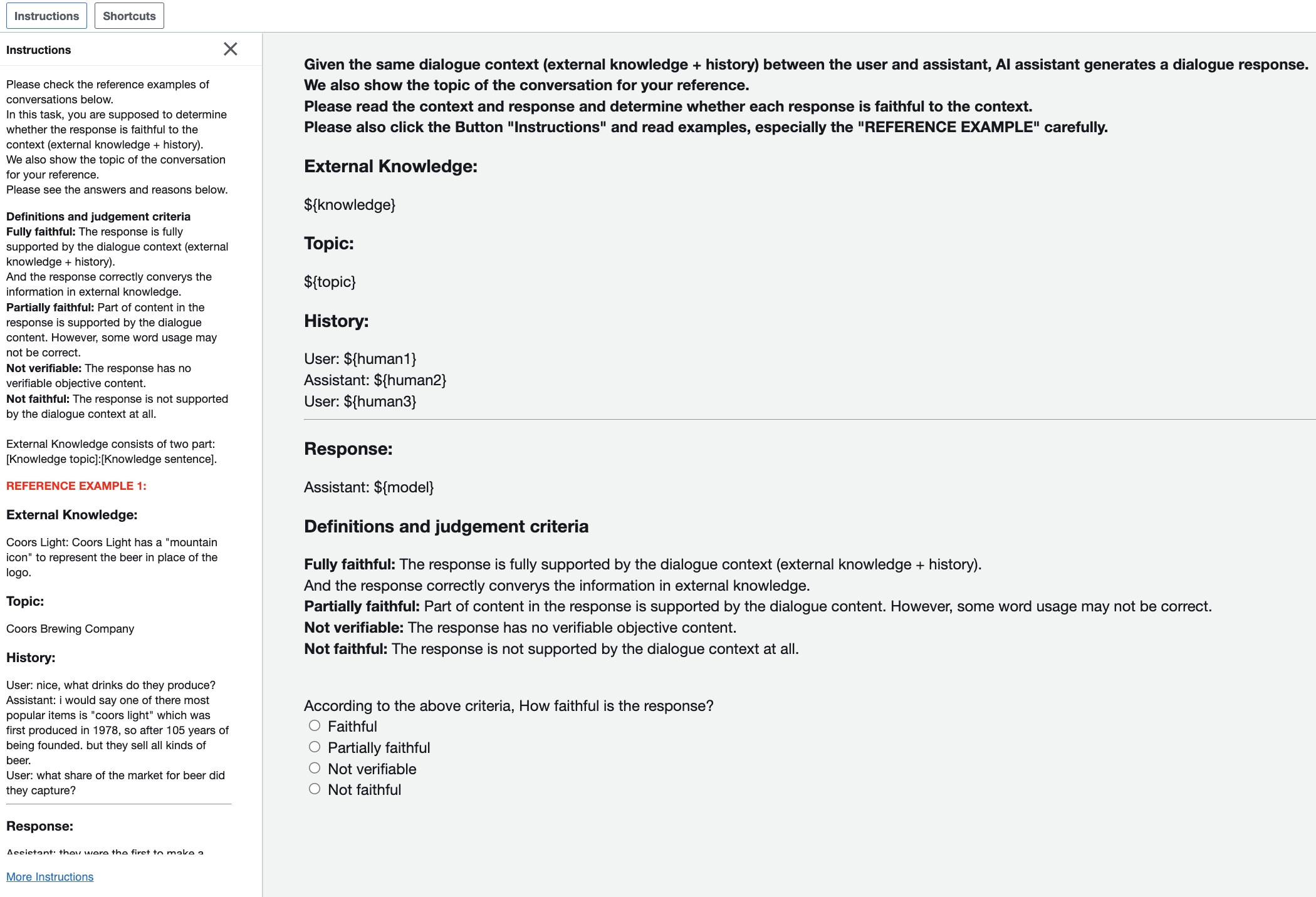}

  \caption{The annotator instructions for human evaluation on \textit{Faithfulness} via 4-point Likert scale.}
  \label{fig:human eval faithfulness}
\end{figure*}

In addition to the automatic evaluation, we conduct human evaluation to assess the quality of responses generated by our model \textit{SPI} and baseline \textit{KnowledGPT} on WoW.
We randomly select 50 samples from each model, and each sample is evaluated by three different annotators. For each comparison, the same context, and two generated responses from each model are shown to the annotators.
We require the annotators to be masters with the following qualifications: the numbers of Human Intelligence Tasks (HITs) approved are greater than or equal to 5000, and their HIT approval rates are greater than or equal to 95\%. The locations of annotators are restricted to Australia, Canada, the United Kingdom, and the United States. 
After collecting annotations for Amazon Mechanical Turk (AMT), we calculate each score as follows: For A/B testing on Fluency and Relevance, we give one score to the model if it generates an equally good or better response than the other one. We present the ratio of the number of data samples with one score over all the test samples. For 4-point Likert scale on Faithfulness, we assign different levels of faithfulness scores from four to one, then present the average score over all the test samples.
\Cref{fig:human eval fluency}, \Cref{fig:human eval relevance}, and \Cref{fig:human eval faithfulness} display the annotator instructions of AMT for \textit{Fluency}, \textit{Relevance}, and \textit{Faithfulness}, respectively. Please find the instructions and examples for annotators in these figures.

\end{document}